\documentclass{article}

\usepackage{microtype}
\usepackage{graphicx}
\usepackage{subfigure}
\usepackage{booktabs} 

\usepackage{hyperref}

\usepackage[accepted]{icml2019}

\icmltitlerunning{A Nonparametric Multi-view Model for Estimating Cell Type-Specific Gene Regulatory Networks}

\usepackage{microtype}

\usepackage{graphicx} 
\usepackage{subfigure} 

\usepackage{natbib}

\usepackage{algorithm}
\usepackage{algorithmic}
\usepackage{amsmath}
\usepackage{amssymb}
\usepackage{hyperref}
\usepackage{color}

\newtheorem{thm}{Theorem}
\newtheorem{lemma}[thm]{Lemma}

\newcommand{\iiid}{\stackrel{\mathrm{iid}}{\sim}}
\newcommand{\ind}{\stackrel{\text{ind}}{\sim}}

\begin{document}

\twocolumn[
\icmltitle{A Nonparametric Multi-view Model for Estimating \\ Cell Type-Specific Gene Regulatory Networks}



\icmlsetsymbol{equal}{*}

\begin{icmlauthorlist}
\icmlauthor{Cassandra Burdziak}{equal,to}
\icmlauthor{Elham Azizi}{equal,to}
\icmlauthor{Sandhya Prabhakaran}{to}
\icmlauthor{Dana Pe'er}{to}
\end{icmlauthorlist}

\icmlaffiliation{to}{Computational \& Systems Biology Program, Memorial Sloan Kettering Cancer Center, New York, NY, USA}

\icmlcorrespondingauthor{Cassandra Burdziak}{cnb3001@med.cornell.edu}
\icmlcorrespondingauthor{Elham Azizi}{mail@elhamazizi.com}

\icmlkeywords{Machine Learning, ICML}

\vskip 0.3in
]



\printAffiliationsAndNotice{\icmlEqualContribution} 

\begin{abstract}
We present a Bayesian hierarchical multi-view mixture model termed \emph{Symphony} that simultaneously learns clusters of cells representing cell types and their underlying gene regulatory networks by integrating data from two views: single-cell gene expression data and paired epigenetic data, which is informative of gene-gene interactions. 
This model improves interpretation of clusters as cell types with similar expression patterns as well as regulatory networks driving expression, by explaining gene-gene covariances with the biological machinery regulating gene expression. 
We show the theoretical advantages of the multi-view learning approach and present a Variational EM inference procedure. We demonstrate superior performance on both synthetic data and real genomic data with subtypes of peripheral blood cells compared to other methods.  
\end{abstract}

\begin{figure*}[hbt]
\centering
\includegraphics[width=0.9\linewidth]{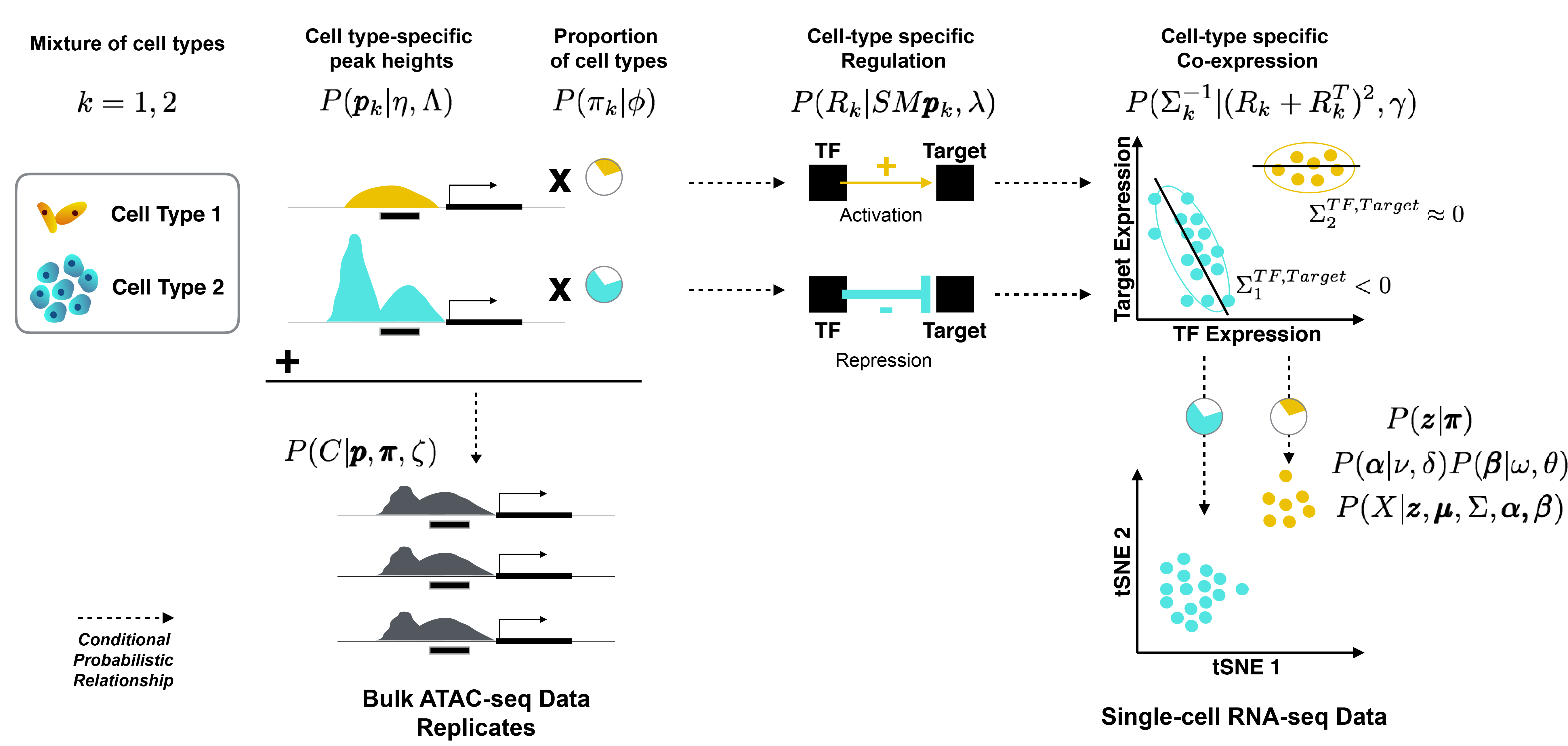}
\caption[]{The generative process in \emph{Symphony}: We aim to infer gene regulatory networks (GRNs) denoted as $R_k$ specific to each cluster of cells from integration of two views:  Epigenetic data (indicative of network edges) from the bulk of cells ($C$) and expression of genes (network nodes) at resolution of single cells ($X$). GRNs are directed weighted networks with edge weights denoting regulatory impact of one gene on another  (e.g. activation or repression); the impact of  regulation is  reflected in covariance between  genes ($\Sigma_k$).} 
\label{fig:generativeprocess}
\vspace{-3mm}
\end{figure*}

\begin{figure*}[hbt]
\centering
\includegraphics[width=0.7\linewidth]{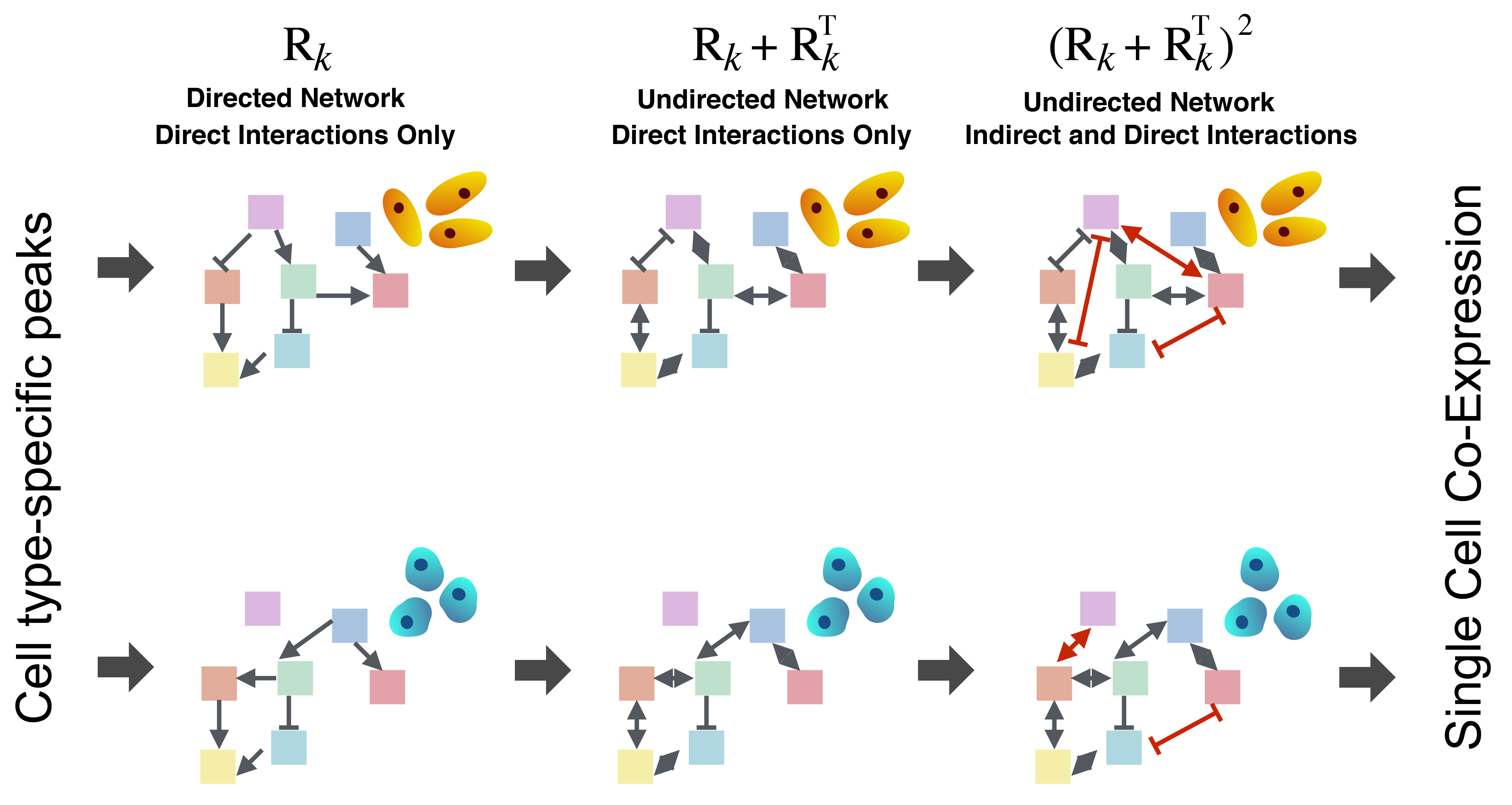}
\caption[]{\emph{Symphony} captures direct and indirect regulation. The impact of  regulation ($R_k$) is propagated through the network up to path length of two and is reflected in covariance between indirectly connected genes ($\Sigma_k$).} 
\label{fig:indirectlinks}
\vspace{-3mm}
\end{figure*}

\section{Introduction}

Joint analysis of different types of data that are associated with the same underlying phenomenon is more informative than  analysis of individual data types, and increases signal to noise ratio \citep{xu2013survey,wang2013multi,rey2012copula}.
This approach, known as multi-view learning or learning with multiple distinct features, has been successfully used in various settings \citep{li2002statistical,jones2003fast,hardoon2004canonical,pan2007adaptive}.

Here, we apply such a multi-view learning approach to address an important biological problem by integrating two views of gene regulation. Our goal is to infer cell clusters (characterizing cell types) as well as their underlying gene regulatory networks (GRNs), which are directed weighted networks between genes depicting the extent to which a regulatory gene influences the expression of each of its downstream {\it target} genes. Understanding differences between regulatory mechanisms across different cell types provides valuable insight in normal development of cell types \citep{davidson2010regulatory}, and mechanisms disrupted in cancer cells 
\citep{pe2011principles,kreeger2009cancer}.



%
Recent advances in single-cell genomic technologies \citep{celseq201,jaitin2014massively,shalek2013single} which measure gene expression at the resolution of individual cells, present remarkable opportunities to characterize different cell types by clustering cells based on heterogeneity of gene expression (as observed features) \citep{satija2015spatial,macosko2015highly}. Learning GRNs from gene expression data alone, however, leads to detection of spurious network links based on correlated genes, while integrative learning from multiple data sources has been shown to improve overall joint inference \citep{zhu2008integrating,hecker2009gene,azizi2014learning}.  
Therefore, we aim to  identify GRNs driving heterogeneous cell types through integrating single-cell expression data with other genomic data types. 
In particular, epigenetic technologies such as ATAC-seq \citep{buenrostro2015atac},  scan the genome for accessible DNA regions, identifying potential interaction between a gene and regulator proteins translated from other genes. In other words, epigenetic data contains information about direct regulatory links between genes and incorporation of epigenetic data is a promising direction for improved inference of GRNs \citep{guo2017chromatin,rotem2015single}. 


We present a novel integrative model, which we refer to as \emph{Symphony}, as a Dirichlet process mixture model that jointly learns clusters of cells and GRNs specific to each cluster. \emph{Symphony} is an extension of the BISCUIT model \citep{prabhakaran2016dirichlet,azizi2018single} which clusters cells while simultaneously distinguishing biological heterogeneity from technical noise in single-cell gene expression data. This is done through incorporating cell-specific parameters scaling the cluster means and covariances for a multivariate Gaussian mixture model.

We extend the BISCUIT model and replace the hyperparameters with a generative process exclusively driven by the paired epigenetic data, which captures the biological mechanism responsible for observed gene covariances per cell type. Briefly, the epigenetic profiles which denote accessible DNA in the bulk samples are deconvolved into cell-type specific accessible regions (Figure \ref{fig:generativeprocess}). Within these regions, the binding of regulatory proteins translated from genes impacts the expression of nearby genes, such that accessible regions may be mapped to gene-gene interactions. This mapping is based on prior knowledge of recurring DNA sequences (known as motifs) associated with these regulatory proteins which occur in regions of accessible DNA. Most importantly, the covariance in observed gene expression is related to a graph power of the regulatory network, capturing the propagated impact of regulation in the network (indirect regulation)(Figure \ref{fig:indirectlinks}).

This  multi-view framework can also be applied to other settings, such as text characterization. For example, to learn the context of queries (vector of words), the bag-of-words simplification may not be sufficient \citep{biemann2005ontology}.
However, the order of words, which can be represented as a latent directed network can imply the context, and incorporating observations from this network such as the frequency of one word following another (as a second view) can enhance extraction of context and clustering of queries (observed as the first view) \citep{landauer1997well,recchia2009more}.   

\begin{figure}[tb]
\centering
\hspace{-10mm}
\includegraphics[width=.7\linewidth]{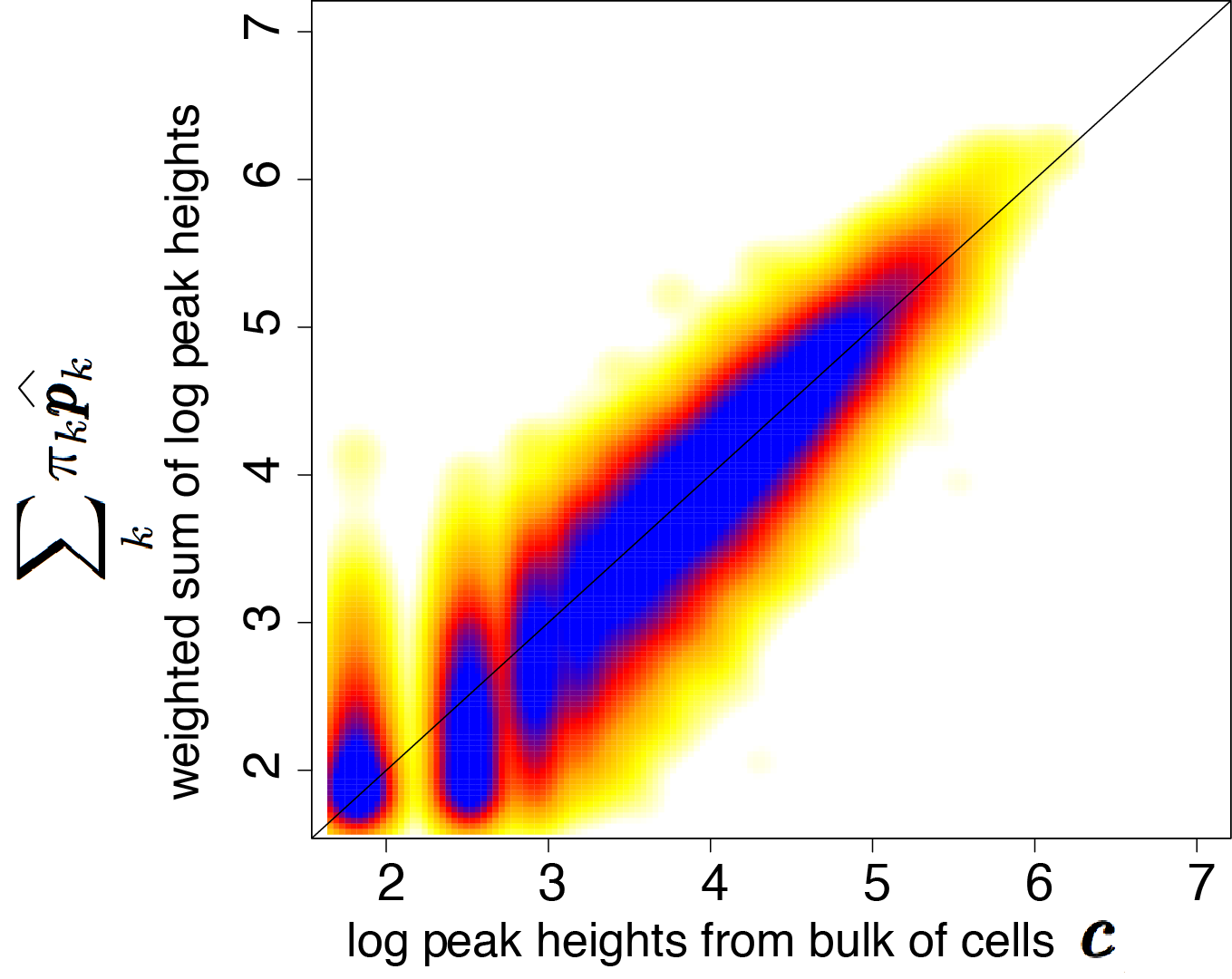}\vspace{-3mm}
\caption[]{Validation of model assumption. Weighted sum of actual peak heights $\hat{\pmb p}_k$s measured from sorted clusters of CD34+ hematopoeitic cells using ATAC-seq, with weights proportional to proportions of cell types, compared to measured peak height from ATAC-seq on the bulk of cells $\pmb c$; heatmap shows density with yellow (low) to blue (high). }\vspace{-3mm}
\label{fig:weightedsum}
\end{figure}

{\bf Related technologies and methods.}
The problem of inferring GRNs specific to cell types involves identifying differences in gene-gene interactions across cell types. 
However, in most cases the cell types are not well-characterized, hence gene markers are not known to enable sorting of cell types prior to measuring epigenetic data (gene-gene interactions). Therefore, epigenetic data measured on the bulk of cells represents a mixture of cell type-specific epigenetic profiles.
One solution is measuring epigenetic profiles at the resolution of single cells. 
These technologies have only recently emerged \citep{buenrostro2015single}; we therefore constructed a model to allow integration of bulk ATAC-seq data. \emph{Symphony} can be easily adapted for inferring GRNs from single-cell ATAC-seq data as well.
%

Other works have attempted to apply computational deconvolution algorithms intended for bulk expression data, such as those using source separation techniques \citep{houseman2016reference}, NMF-based methods \citep{repsilber2010biomarker} or Bayesian models \citep{erkkila2010probabilistic}, to instead infer cell type-specific  epigenetic profiles. 
Recent methods such as SCENIC \citep{aibar2017scenic} infer GRNs from single-cell expression data alone and do not incorporate epigenetic or other types of data.

{\bf Contributions.}
 In this paper, we show that a multi-view learning framework would improve the deconvolution of epigenetic data.
  Furthermore, using an integrative model, we improve  the clustering of cells, and hence characterization of cell types. 
Most importantly, our model presents the advantage of inferring cell type-specific GRNs that give insight into heteroegeneity of underlying mechanisms across cell types.  We present a Variational EM inference procedure and show that the integration guarantees model identifiability, while learning from the epigenetic view alone does not.
 While other works have attempted to integrate bulk multi-omics data \citep{lake2017integrative,brown2013integrative,ritchie2015methods}, there are no methods to our knowledge that infer heterogeneous GRNs through integrating epigenetic and single-cell resolution gene expression data.
 %


\begin{figure}[t]
\centering
\includegraphics[width=.9\linewidth]{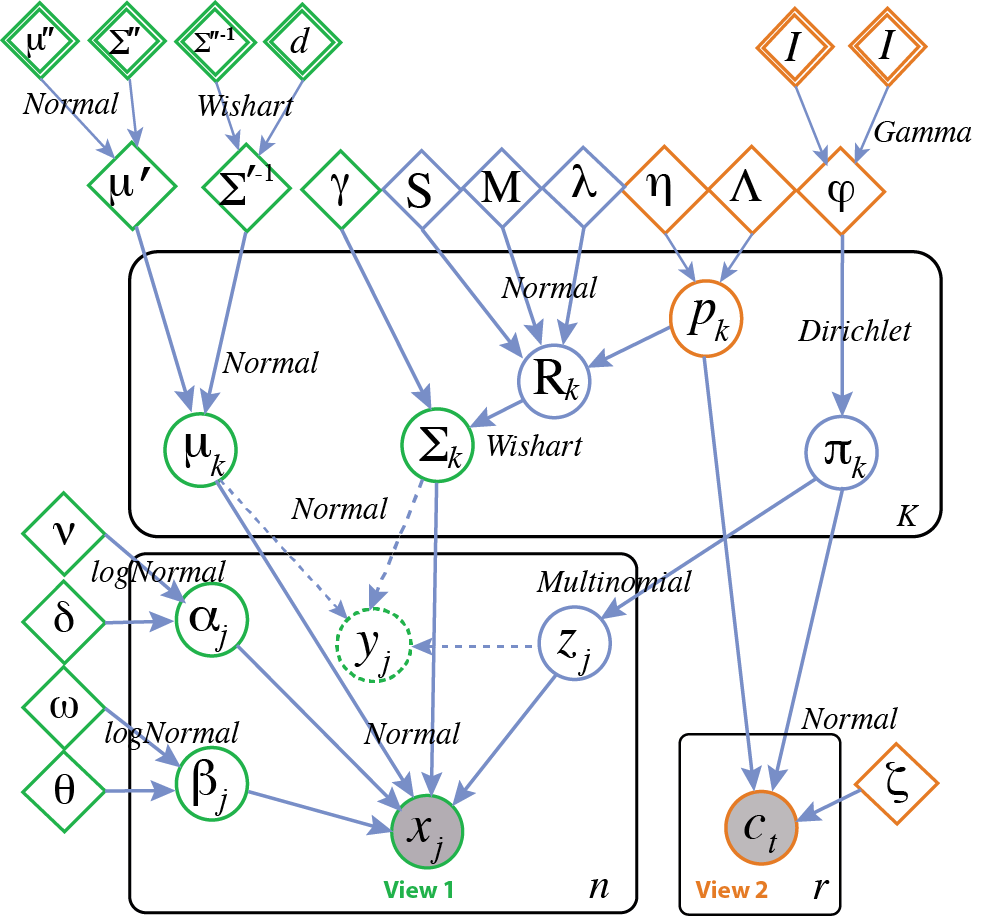}
\caption[]{Plate model for \emph{Symphony}. white circles denote
latent variables of interest,  diamonds are hyperparameters and double diamonds are hyperpriors calculated empirically. }\vspace{-5mm}
\label{fig:Plate_model}
\end{figure}

\section{Model}
The observed data is considered as two views from the biological system (Figure \ref{fig:generativeprocess}): \vspace{-1mm}

\pmb{View 1.} Single-cell gene expression data from scRNA-seq technologies \citep{klein2015droplet,macosko2015highly} denoted as $X^{d \times n}=[ \pmb{x}_1,\cdots,\pmb{x}_j,\cdots,\pmb{x}_n ]$ where each observation $\pmb{x}_{j} \in \mathbb{R}^d$ for cell $j \in \{1,\cdots,n\}$
corresponds to $d$ genes (as features). 
Each entry $x_{ij}$ for $i=[1,\cdots,d]$ contains the expression of gene $i$ in cell $j$ (more precisely, the log of counts of mRNA molecules per gene $i$ from cell $j$ plus a pseudo-count). \vspace{-1mm}

\pmb{View 2.} Epigenetic data, for example measured with ATAC-seq technology \citep{buenrostro2015atac}, denoted as $C^{l \times r}=[\pmb{c}_1,  \cdots, \pmb{c}_t, \cdots, \pmb{c}_r]$ where each observation $\pmb{c}_t \in \mathbb{R}^l$ for $t \in \{1,\cdots,r\}$  corresponds to $l$ genomic regions (as features). Specifically, $\pmb{c}_t$ is an experimental replicate measuring accessibility of genomic regions $m=[1,\cdots,l]$. \vspace{-1mm}

{\bf Prior knowledge.} The genomic regions in $C$ can be mapped to genes in $X$ with a pre-defined mapping function $g(i,i^\prime)=m$ that relates each genomic region $m \in \{1,\cdots,l \} $ to a gene-gene interaction $i^\prime \rightarrow i$ for $i,i^\prime \in \{1,\cdots,d \}$. 
%
We also define $M^{d \times d}$ based on prior knowledge containing binary values $M_{i,i^\prime} =1$ if the motif sequence for gene $i^\prime$ exists in the genomic region $m$ in the vicinity of gene $i$, meaning a potential interaction can exist from gene $i^\prime$ to gene $i$. 

\subsection{Epigenetic Model (View 2)}
\vspace{-2mm}
The epigenetic data is informative of network structure, i.e. existence of edges between genes (features) and gene expression data contains information on network nodes.  
We aim to infer this directed weighted network (GRN) for each cluster $k \in \{1,\cdots,K\}$ of cells, denoted by the asymmetric matrix $R_k^{d \times d}$ in which entry $R_{k_{i,i^\prime}} \neq 0$ if $i^{\prime} \rightarrow i$, meaning gene $i^\prime$ directly regulates gene $i$.
$R_{k_{i,i^\prime}}$ is the regulatory function of gene $i^\prime$ on gene $i$ in cluster $k$ such that $R_{k_{i,i^\prime}}>0$ or $R_{k_{i,i^\prime}}<0$ represent activation or repression of expression respectively, with $|R_{k_{i,i^\prime}}|$ being the strength of regulation. 

We do not aim to distinguish all layers of the regulatory process (such as protein phosphorylation) and rather interpret GRNs as an approximation for the overall impact of TFs on target genes at the transcriptional level.

%

We model regulation of gene expression as follows:
Genome accessibility in cluster $k$ is represented with latent variable ${\pmb p}_k = [p_{k}^1,\cdots, p_{k}^l]^T \in {\mathbb{R}^+}^l$ containing $l$ genomic regions (features). This represents log of peak heights plus 1 (to ensure positive domain) in all genomic regions, for each cell type $k$. We set a truncated multivariate Normal prior to capture the structure between genomic regions encompassing co-regulated genes (i.e. genes sharing regulators) with mean $\pmb \eta$ and covariance $\Lambda$: %
%
$\quad \pmb p_{k} \sim trunc\mathcal{N}(\pmb \eta,\Lambda, \pmb 0, +\infty ). \quad$
In this paper, we assume a setting where we do not observe $\pmb p_k$s, and only observe epigenetic data from the bulk of cells which can be represented as a weighted sum of cluster-specific epigenetic profiles where $\pi_k$s are weights. 
%
Thus, our epigenetic model is: \vspace{-2mm}
\begin{equation}
\begin{split}
\{\pmb c\}_{t}^{(1,\cdots,l)}|{\pmb p}_k, \pi_k &\ind \mathcal{N}(\sum_k \pi_k {\pmb p}_k, \zeta I) 
\end{split}
\end{equation}
\vspace{-8mm}

We validated the above assumption of weighted sum  using ATAC-seq data from hematopoeitic progenitor cells from \citet{corces2016lineage} by computing the weighted sum of measurements from sorted cell types (actual $\pmb p_k$s denoted with $\hat{\pmb p}_k$) to measurements from the bulk of cells as $\pmb c$ (Figure \ref{fig:weightedsum}).

We have a  $K$-order Dirichlet prior over $\pi_k$: 
$\pi_k| \varphi,K \sim \mathit{Dir} ( \pi_k|\tfrac{\varphi}{K}, \cdots, \tfrac{\varphi}{K}) $, where 
$\varphi^{-1} \sim \mathit{Gamma}(1,1)$. 
%
%
Then, in cell type $k$, if a genomic region $m$ in the vicinity of gene $i$ is accessible with log peak height $p_k^m$, and the motif sequence associated with one or more Transcription Factor (TF) proteins translated from genes $i^\prime,i^{\prime\prime},...$ exists in the region ($M_{i,i^\prime} =M_{i,i^{\prime\prime}} =1$), then the TF(s) can bind to the region and hence regulate the expression of gene $i$. 
Furthermore, we assume the peak height (strength of TF binding to genome) is informative of $i^\prime \rightarrow i$ edge weight $|R_k|$ (strength of regulation). Thus, we model $R_k$ as follows: \vspace{-2mm}
\begin{equation}
\begin{split}
R_k^{i,i^\prime} &\sim \mathcal{N} ( S^{i,i^\prime} M^{i,i^\prime}  \pmb p_k^{g(i,i^\prime)}, \lambda)
\end{split}
\label{eq:modelR}
\end{equation}
 The function $g(\cdot)$ maps gene pair $i,i^\prime$ to genomic region $l$. $S$ denotes a sign indicator variable representing repression or activation function. We set $S$ according to the sign of the empirical covariance:  $S^{i,i^\prime}=sign(\Sigma^{\prime\prime^{i,i^\prime}})$.
%

\subsection{ Single-cell Gene Expression Model (View 1)}\vspace{-2mm}
We use the above epigenetic model to drive gene expression data based on the following key ideas: First, if a direct regulatory link exists from a TF associated with gene $i^\prime$ to a target gene $i$ ($i^{\prime} \rightarrow i$), we assume there is strong covariance between their expressions. 
Second, due to van der Corput's inequality \citep{montgomery2001harmonic}, covariances can partly reflect the propagated impact of indirect regulation in cases where genes are not directly connected but exist on the same path in the network (Figure \ref{fig:indirectlinks}).
%
%
For example if $i^{\prime\prime} \rightarrow i^{\prime}$ and $i^{\prime} \rightarrow i$, we might also observe covariance between $i,i^{\prime\prime}$ even though they are not directly connected in the network (e.g. $R_k^{i,i^{\prime\prime}} = 0, \Sigma_k^{i,i^{\prime\prime}} \neq 0$).
%
Here, we consider indirect effects with path length up to two using the square of the indirected network $(R_k+R_k^T)^2$ \citep{walker1992implementing} such that:
\begin{equation}
\begin{split}
\Sigma_k^{-1}| R_k & \sim  \mathit{Wish}({(R_k+R_k^T)}^{-2}, \gamma) \\
\end{split}
\label{eq:modelSigma}
\end{equation}
${(R_k+R_k^T)}^2$ is positive semi-definite according to Lemma 3 in the following section making the above modeling assumption feasible. Additionally, this model can capture combinatorial regulation in the inferred covariances. In particular, a gene pair $i,i^\prime$ will always have the same directionality of regulation (i.e. activation or repression relationship), but  $\Sigma_k^{i,i^\prime}$ can be positive in one cluster and negative in another cluster depending on the relative regulatory strength of activators and regulators in its path. An example of this variability in sign is shown in Supplmentary Figure \ref{fig:covsim}.

Gene expression data for each cell $j$ denoted as $\pmb x_j$ is then modeled similar to the multivariate Gaussian mixture model in BISCUIT:\vspace{-2mm}
\begin{equation}
\begin{split}
\{\pmb x\}_{j}^{(1,\cdots,d)}|z_j=k \ind & \mathcal{N}(\alpha_j\pmb\mu_k, \beta_j\Sigma_k) \\
\pmb \mu_k \sim \mathcal{N}(\pmb \mu^\prime,\Sigma^\prime), &  \quad
\pmb \mu^\prime \sim \mathcal{N}(\pmb \mu^{\prime\prime},\Sigma^{\prime\prime})\\
\Sigma^{\prime -1} \sim \mathit{Wish}(d,\frac{1}{d \Sigma^{\prime\prime}}), & \quad
z_j|\pi_k  \iiid \mathit{Mult}(z_j| \pi_k) \\
\alpha_j \sim log\mathcal{N}(\nu, \delta^2) , & \quad
\beta_j \sim log\mathcal{N}(\omega, \theta) 
\end{split}
\label{eq:expmodel}
\end{equation}

\vspace{-2mm}
where $z_j$ denotes assignment of cell $j$ to cluster $k \in \{1,\cdots,K\}$.
With integrating two data modalities (gene expression and epigenetic data), we improve inference of clusters (cell types). 
The scaling parameters $\alpha_j,\beta_j$ specific to cell $j$ are used to normalize the data $\pmb x_j$ in downstream analysis by transforming to $\pmb y_j \sim \mathcal{N}(\pmb \mu_k,\Sigma_k)$ according to the cluster it is assigned to $z_j=k$, similar to \citet{prabhakaran2016dirichlet}.
%
The plate model for \emph{Symphony} is summarized in Figure \ref{fig:Plate_model}.

\section{Theory}
We show theoretical advantages of integration of the two data types using \emph{Symphony} as follows: 
We define $f(\pmb x):=\mathcal{N}(\alpha\pmb\mu_k, \beta\Sigma_k) \in \mathbb{R}^d$ as the multivariate Gaussian density of $\pmb x$ and $f(\pmb c):=\mathcal{N}( \sum_k \pi_k {\pmb p}_k, \zeta I) \in \mathbb{R}^l$ as the multivariate Gaussian density of $\pmb c$. 
First, we emphasize that the epigenetic model alone $f(\pmb c|\pmb{p}_k, \pi_k, \zeta)$ is not identifiable and therefore precise inference of  deconvolved epigenetic profiles ($\pmb{p}_k$s) is not possible:\vspace{-1mm}
\begin{lemma} The epigenetic model $f(\pmb c|\pmb{p}_k, \pi_k)$ is non-identifiable (Proof in  Supplementary section B) 
\end{lemma} \label{lemdeconv}  \vspace{-2mm}
%
%
 This motivated us to build an integrative model. 
Identifiability of the single-cell expression model $f(\pmb x|\pmb \mu_k, \Sigma_k, \pmb \alpha, \pmb \beta, \pmb z)$ has been shown  under certain enforced constraints on both   $\pmb{\alpha,\beta}$:\vspace{-1mm}

\begin{lemma} \citep{prabhakaran2016dirichlet}
Defining  $\Phi = \{\forall j, k: (\alpha_j,\pmb\mu_k, \beta_j,\Sigma_k)\} \cup \{\pmb\pi\}$
 $\Phi = \Phi^\star$ if the following conditions hold: $\forall j: \pmb\mu_k \ge \pmb\mu^\prime + diag(\Sigma^\prime)(\alpha_j-\nu)/\delta$ and $\forall j: \beta_j \ge \frac{\theta}{\omega+1}$  
\end{lemma}\label{lembiscuit}\vspace{-2mm}
While the above conditions guarantee identifiability, they are not inferred from data or biologically motivated and hence interpretation of parameters may not provide the best characterization of cell types.
Here, we show that in the integrative model, the constraints for $\beta_j$s are no longer required and identifiability of the expression model $f(\pmb x|\pmb \mu_k, \Sigma_k, \pmb \alpha, \pmb \beta, \pmb z, \pmb p_k, R_k)$ is guaranteed through the extension of the model that captures regulation, from which we observe additional epigenetic data $\pmb{c}$. Hence the full model $f(X,C|\{ \pmb \mu_k, \Sigma_k, \pmb \alpha, \pmb \beta, \pmb z, \pmb p_k, R_k, \pi_k, \zeta\})$ is identifiable.
%
%
%
We will use the following lemma  to show $(R_k+R_k^T)^2$ is positive semi-definite.\vspace{-1mm}
\vspace{-1mm}
\begin{lemma}
Square of a symmetric matrix $H$ gives a symmetric positive semi-definite matrix $L$ (Proof in  Supplementary section B).
\end{lemma} \label{lemsympsd}
\vspace{-3mm}

Since single-cell expression data $ X$ usually contains expression of genes that do not have observations in their mapped genomic regions in $\pmb c$ such that $d>l$, we first define a reduced version of the model  where all genes in $\pmb x$ do have mapped genomic regions: $f (X^0,C)$ where $X^0$ is a ${l \times n}$ subset of $X$.  Then, $f(\pmb x^0_{l \times 1})=\mathcal{N}(\alpha \pmb \mu_k^0,\beta \Sigma_k^0)$ with $\pmb \mu_k^0,\Sigma_k^0$ being subsets of $\pmb \mu$ and $\Sigma$. We next show the identifiability of the reduced model. Then, we use this result to extend the identifiability to the full model given $\pmb \beta$ which is the parameter scaling $\Sigma_k$. Finally, we show the identifiability of the full model.

\vspace{-2mm}


\begin{lemma}
In the reduced model: $f(X^0,C|\pmb \beta, \pmb \mu_k^0, \Sigma_k^0, \pmb \alpha, \pmb z, \pmb p_k, R_k^0, \pi_k, \zeta)$, $\beta$s are identifiable under the conditions of: $\forall j: \pmb\mu_k \ge \pmb\mu^\prime + diag(\Sigma^\prime)(\alpha_j-\nu)/\delta$ without the need for condition on $\beta$s (Proof in  Supplementary section B). 
\end{lemma} \label{lembeta}\vspace{-5mm}


\begin{lemma}
For a given $\pmb \beta = \beta^*$, identifiability of: $f(X,C|\pmb \mu_k, \Sigma_k, \pmb \alpha, \pmb \beta=\pmb \beta^*, \pmb z, \pmb p_k, R_k, \pi_k, \zeta)$ is guaranteed if $ \forall j: \pmb\mu_k \ge \pmb\mu^\prime + diag(\Sigma^\prime)(\alpha_j-\nu)/\delta$ (Proof in  Supplementary section B). 
\end{lemma}\vspace{-5mm}
%

\begin{thm}
The full model $f(X,C|\{ \pmb \mu_k, \Sigma_k, \pmb \alpha, \pmb \beta, \pmb z, \pmb p_k, R_k, \pi_k, \zeta\}))$ is identifiable if $ \ \forall j: \pmb\mu_k \ge \pmb\mu^\prime + diag(\Sigma^\prime)(\alpha_j-\nu)/\delta$ (Proof in  Supplementary section B).
\end{thm} 

%




%
\begin{figure*}[tbh]
\centering
\includegraphics[width=1\linewidth]{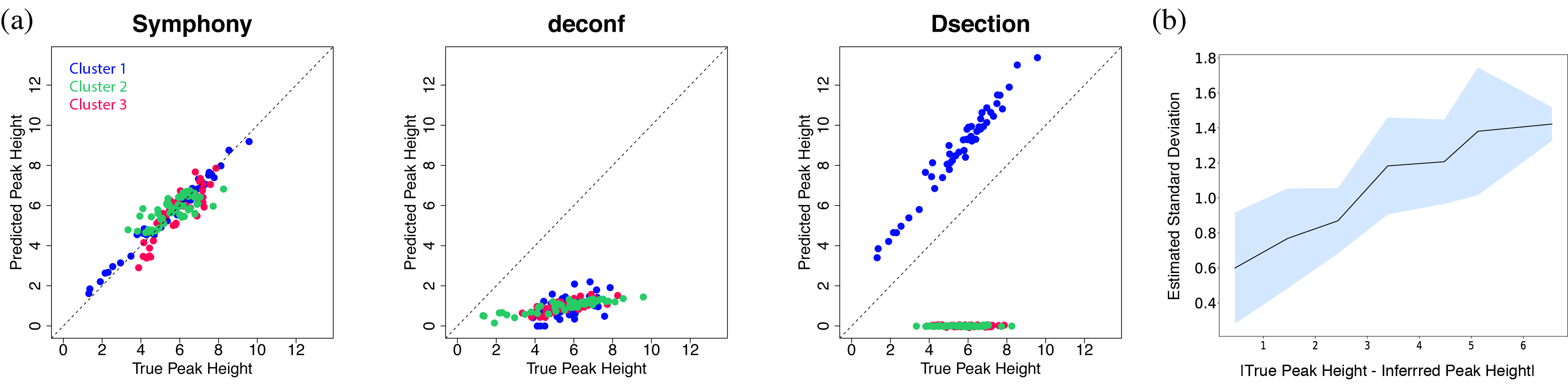}\vspace{-5mm}
\caption[]{Performance in deconvolving epigenetic data. \textbf{(a)} Estimated peak heights $\pmb p_k$ using \emph{Symphony} for $K=3$ synthetic clusters versus true peak heights, compared to two other deconvolution methods: deconf \citep{repsilber2010biomarker} and Dsection \citep{erkkila2010probabilistic}; each dot represents a genomic region. \textbf{(b)} Moving average of estimated standard deviation for $\pmb p_k$s using \emph{Symphony} vs. estimate residual; shaded area shows 1 standard deviation in each window of length 1.} 
\label{fig:pscatter}
\end{figure*}


\begin{figure*}[bht]
\centering
\includegraphics[width=.25\linewidth]{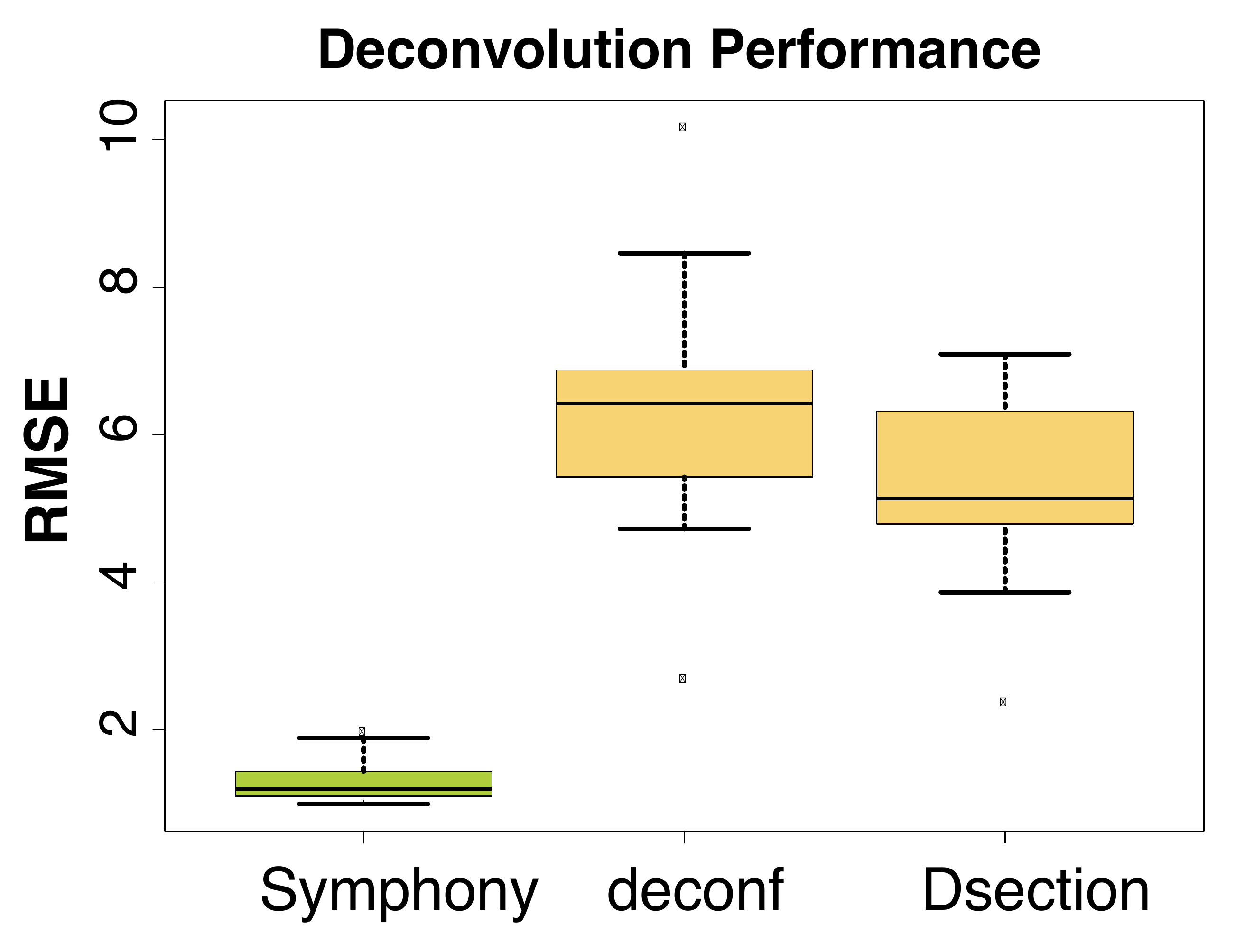}
\includegraphics[width=.37\linewidth]{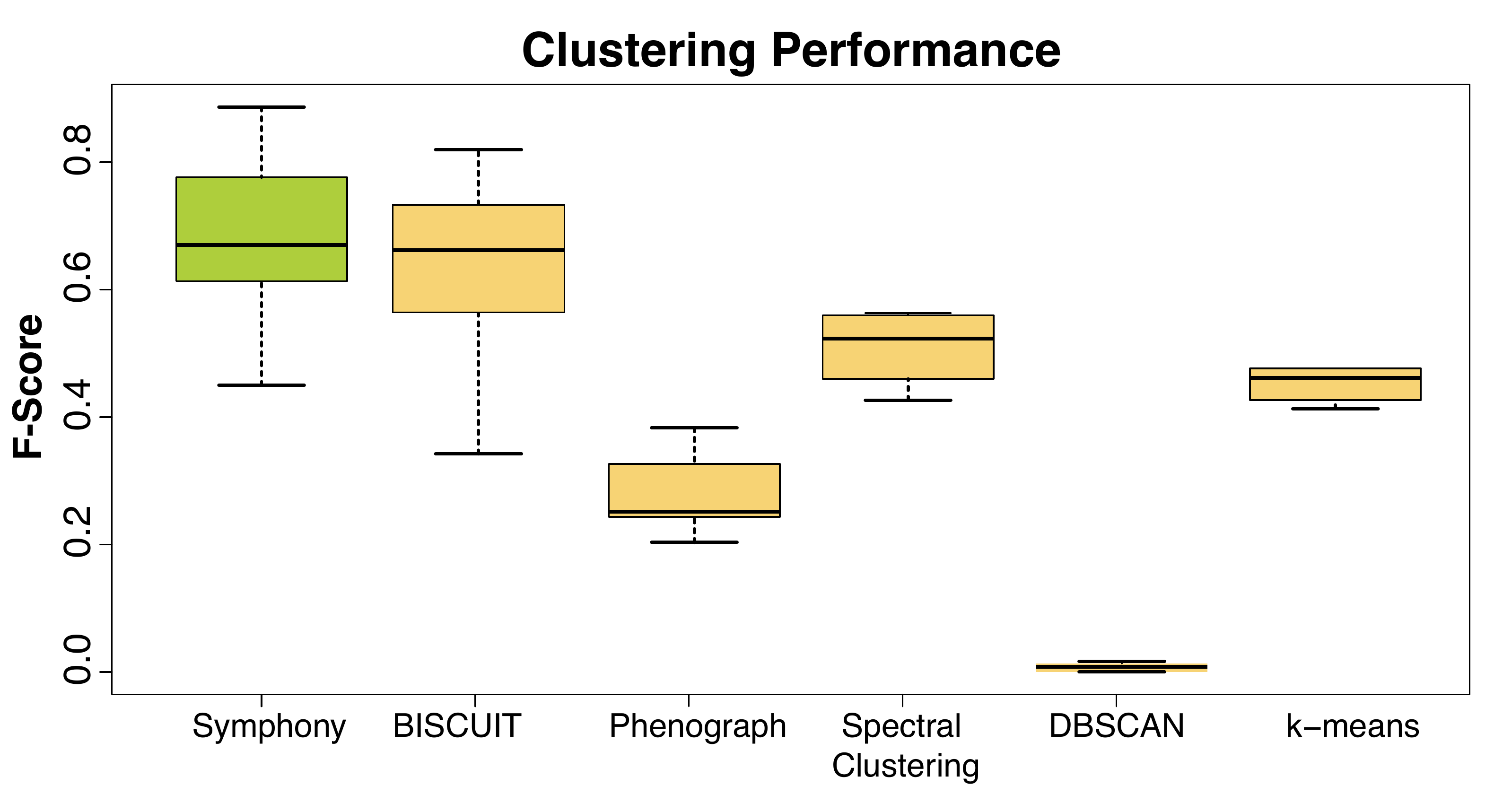}
\includegraphics[width=.36\linewidth]{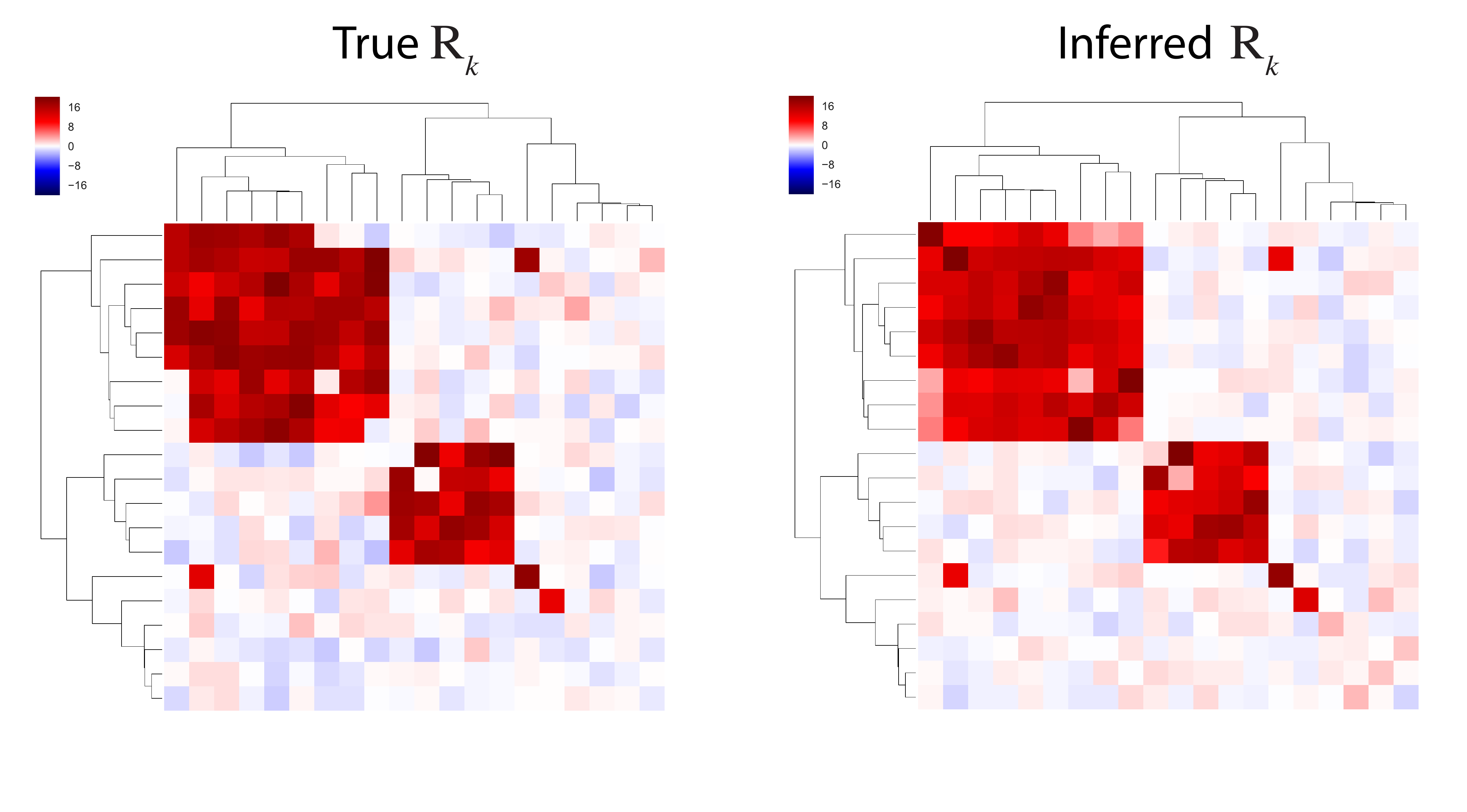}
\caption[]{ {\bf Left:} Root-mean-square error (RMSE) in inferring $\pmb p_k$s across $10$ experiments  compared to other deconvolution methods  used for genomic data. {\bf Middle:} Performance in clustering cells across $10$ experiments compared to other clustering methods commonly used for single-cell gene expression data. {\bf Right:} Heatmap depicting true versus inferred $R_k$ in synthetic data showing \emph{Symphony's} superior capabilities in recovering underlying $R_k$.  }
\label{fig:pboxplot}
\end{figure*}

\begin{figure*}[ht]
\centering
\includegraphics[width=1.12\linewidth]{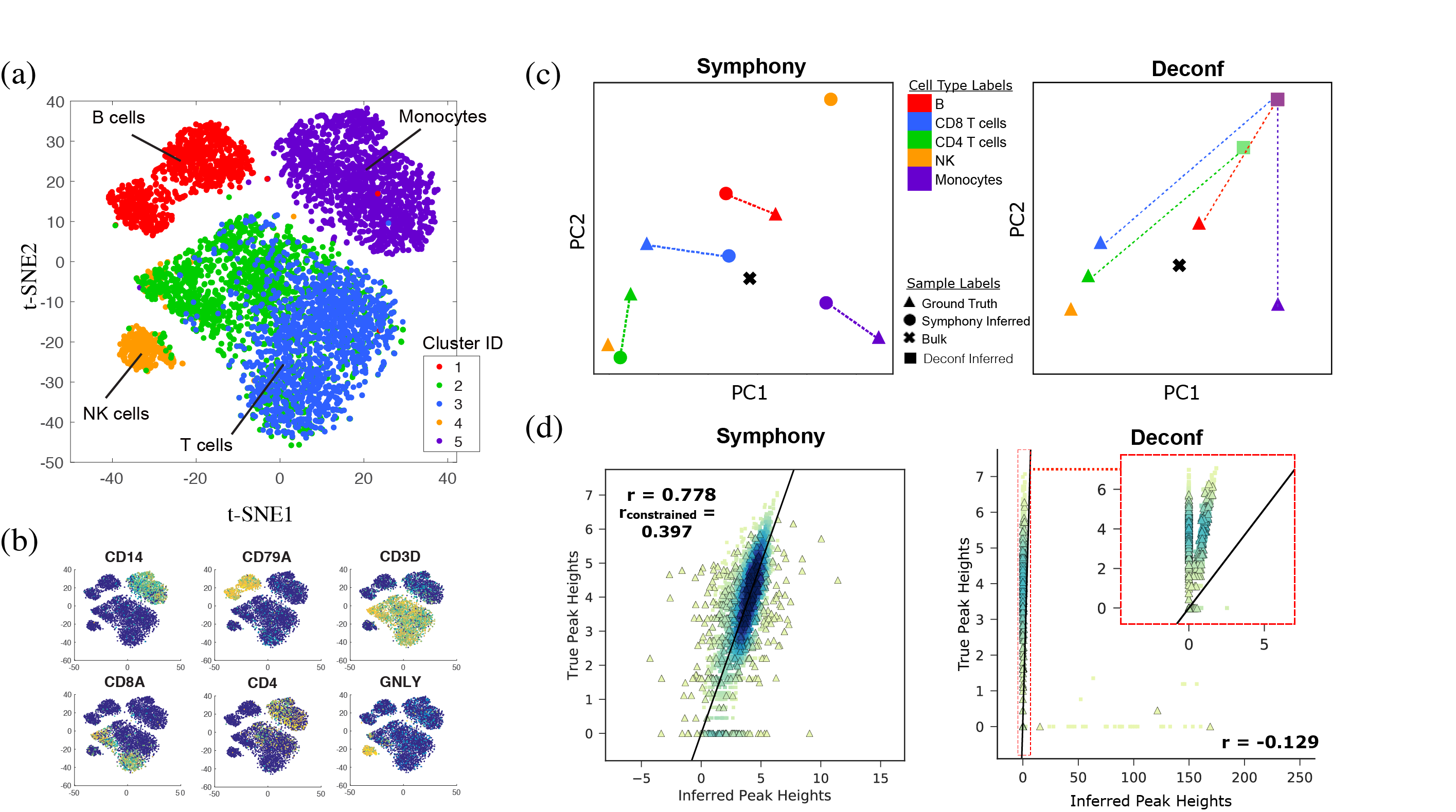}
\caption[]{Performance on genomic data from PBMCs. \textbf{(a)} t-SNE map of $6825$ single-cell gene expression data from 5 cell types after normalization, cells colored by clusters \textbf{(b)} expression of cell-type   markers \textbf{(c)} PCA on expression-constrained peaks and those which show at least some accessibility in all cell types (to show effective peak \emph{magnitude} estimation) showing global performance of Symphony in deconvolving epigenetic data; shown with projection of inferred peak heights on principal components of ground truth peak heights (from ATAC-seq on sorted cell types in \citet{corces2016lineage}) (left) compared to deconvolution using Deconf (right) which fails to deconvolve the majority of peaks shown with overlapping squares \textbf{(d)} Scatterplot of inferred peak heights for all clusters vs ground truth peak height using Symphony (left) compared to Deconf (right); peaks are colored by density; $r$ values show Pearson correlation; peaks constrained by expression data and bulk epigenetic data are triangular points with black outline; NK cells were not included in this plot due to the small cell proportion ($<$ 5\%), making deconvolution impossible. }
\label{fig:pbmc}
\vspace{-3mm}
\end{figure*}

\begin{figure}[hbt]
\centering
\includegraphics[width=1\linewidth]{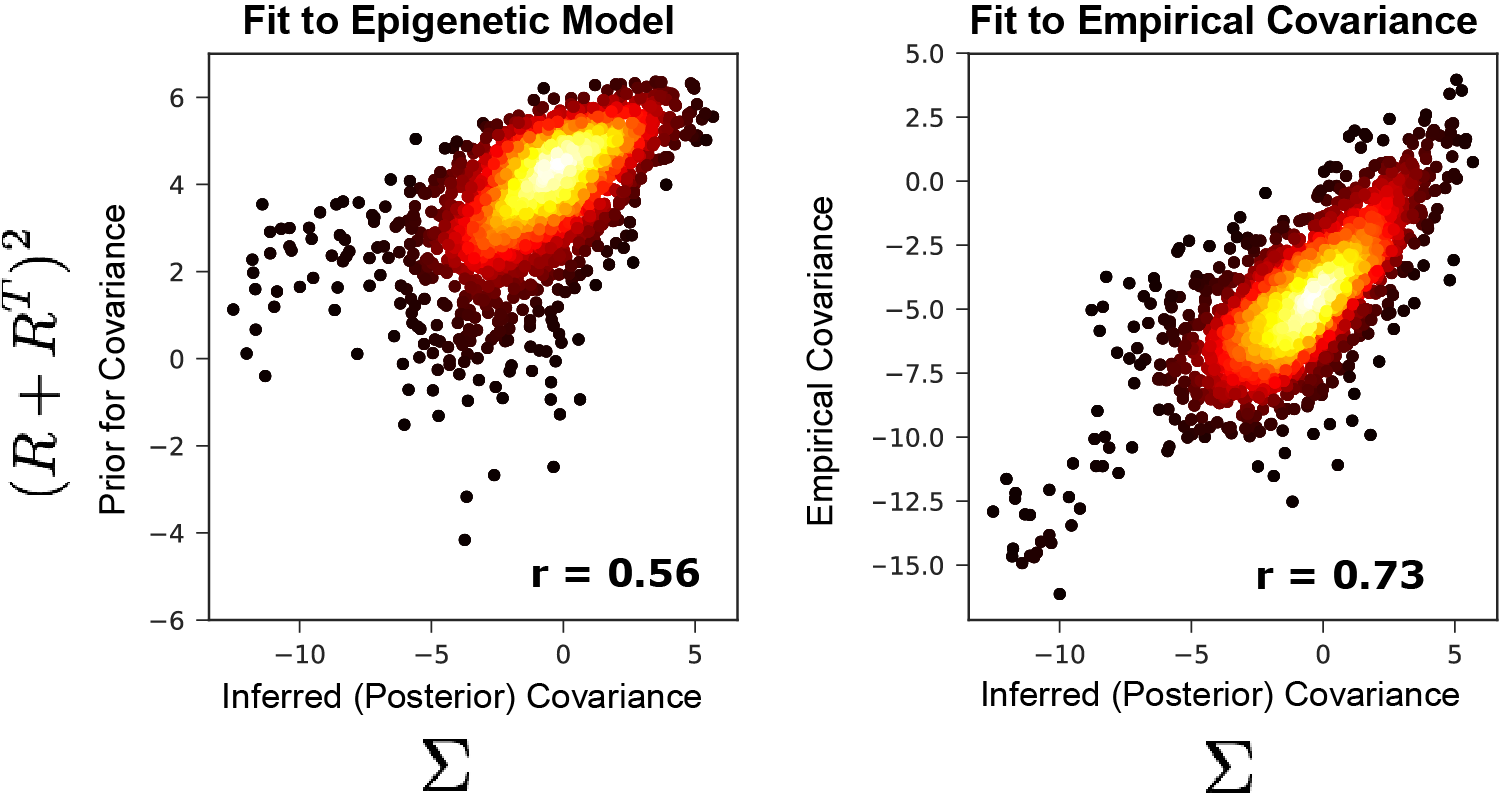}
\caption[]{\emph{Symphony} model fit. \textbf{Left:} Inferred covariance vs its prior capturing direct and indirect regulation $(R+R^T)^2$; \textbf{Right:} Empirical gene covariance compared to inferred covariance across all cell types. All axes are log-transformed. } 
\label{fig:fitprior}
\end{figure}

\begin{figure}[hbt]
\centering
\includegraphics[width=1\linewidth]{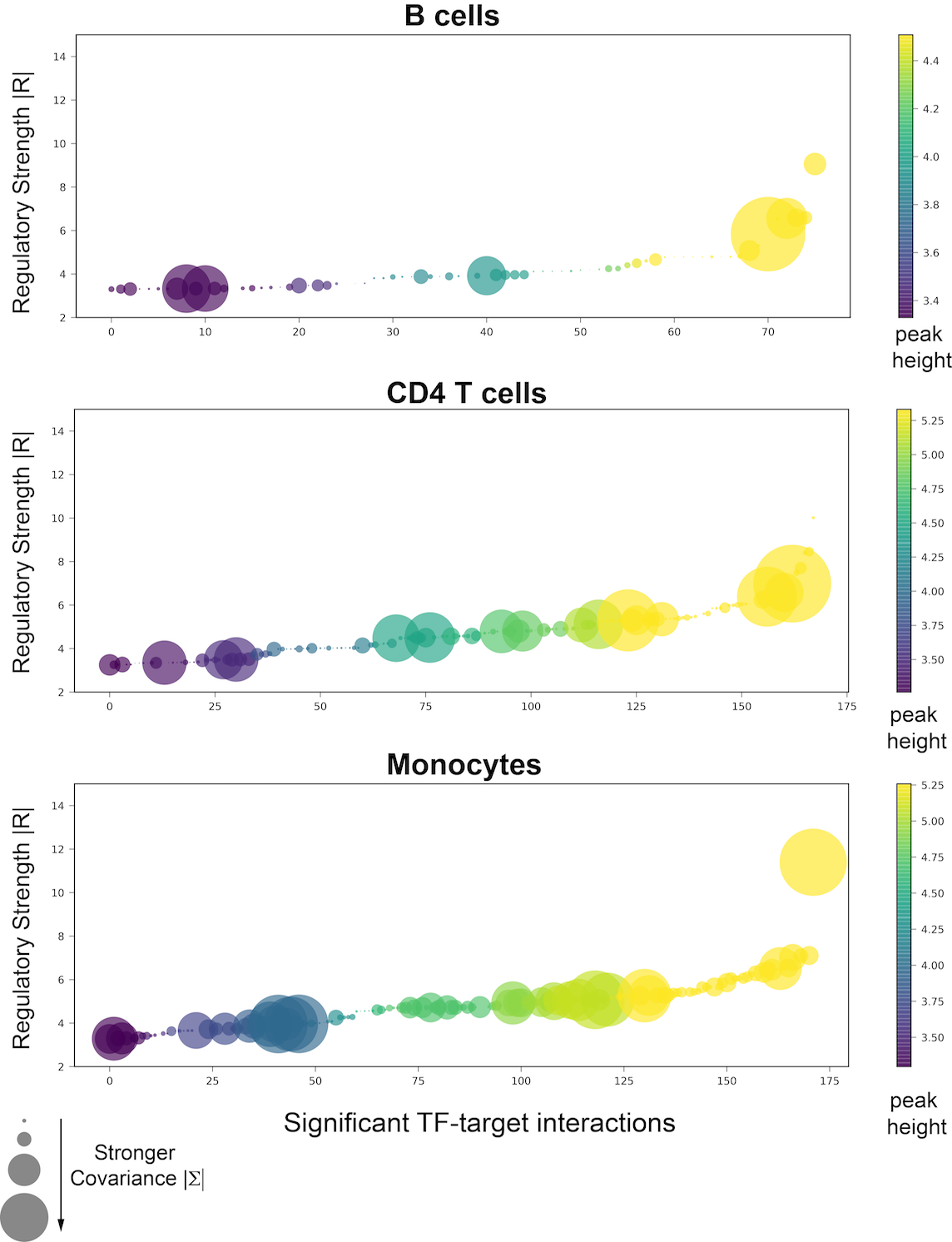} 
\caption[]{GRN Interpretation. Inferred regulations sorted by strength ($|R|$) show association with either peak heights or TF-target covariance or both; circle sizes are proportional to covariance strength ($|\Sigma|$) and they are colored by inferred peak height per cell type. Covariances are z-normalized for scale.}
\label{fig:sortedR}
\end{figure}

\begin{figure*}[htb]
\centering
\includegraphics[width=0.6\linewidth]{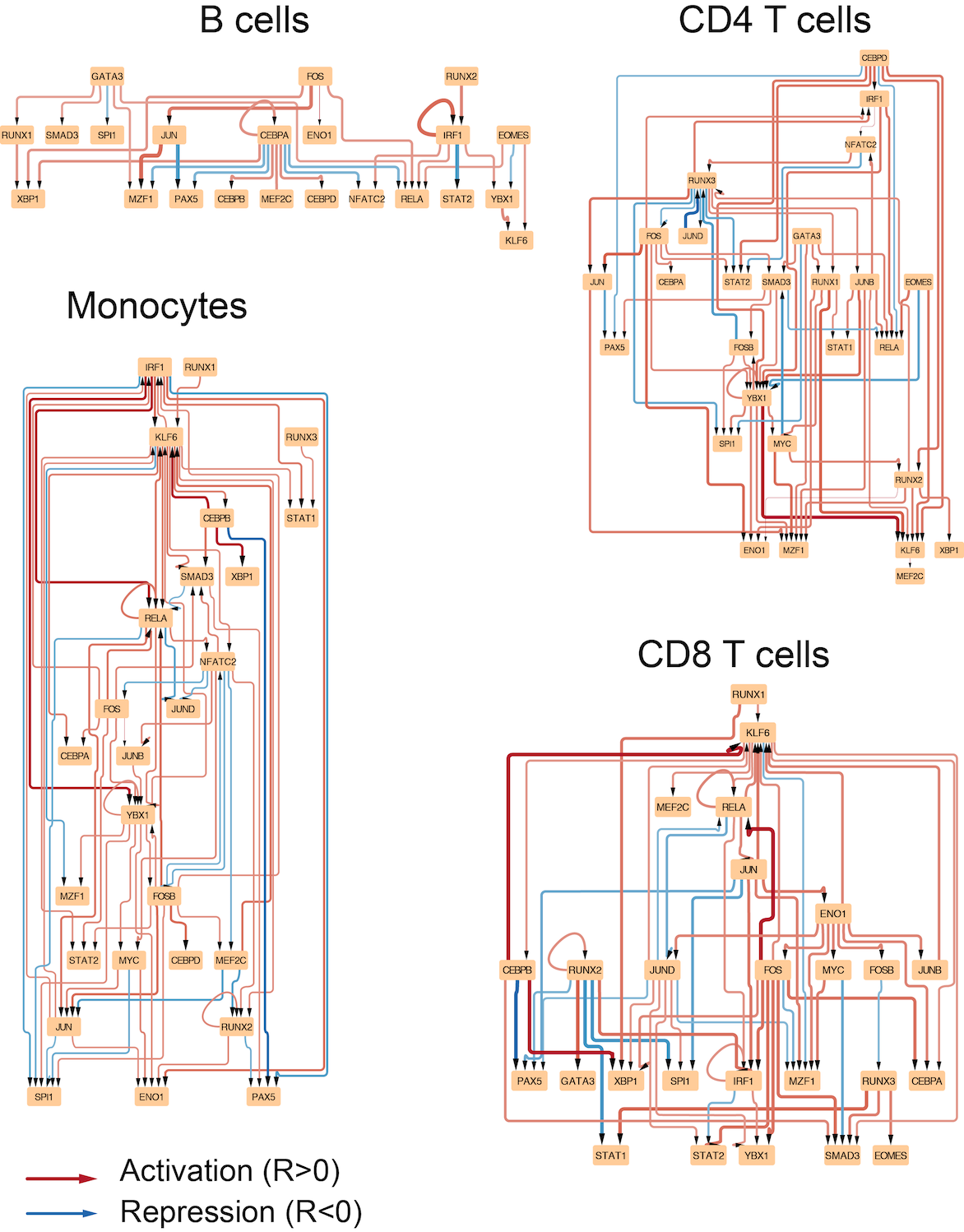} 
\caption[]{Subnetworks of cell type-specific GRNs between TFs with strongest regulations ($|R_k|>4.5$); global structures show differences in connectivity of TFs across clusters (cell types); red and blue edges indicate activation and repression; edge widths are proportional to strength of regulation ($|R|$). 
} 
\label{fig:networks}
\end{figure*}

\begin{figure*}[hbt]
\centering
\includegraphics[width=1\linewidth]{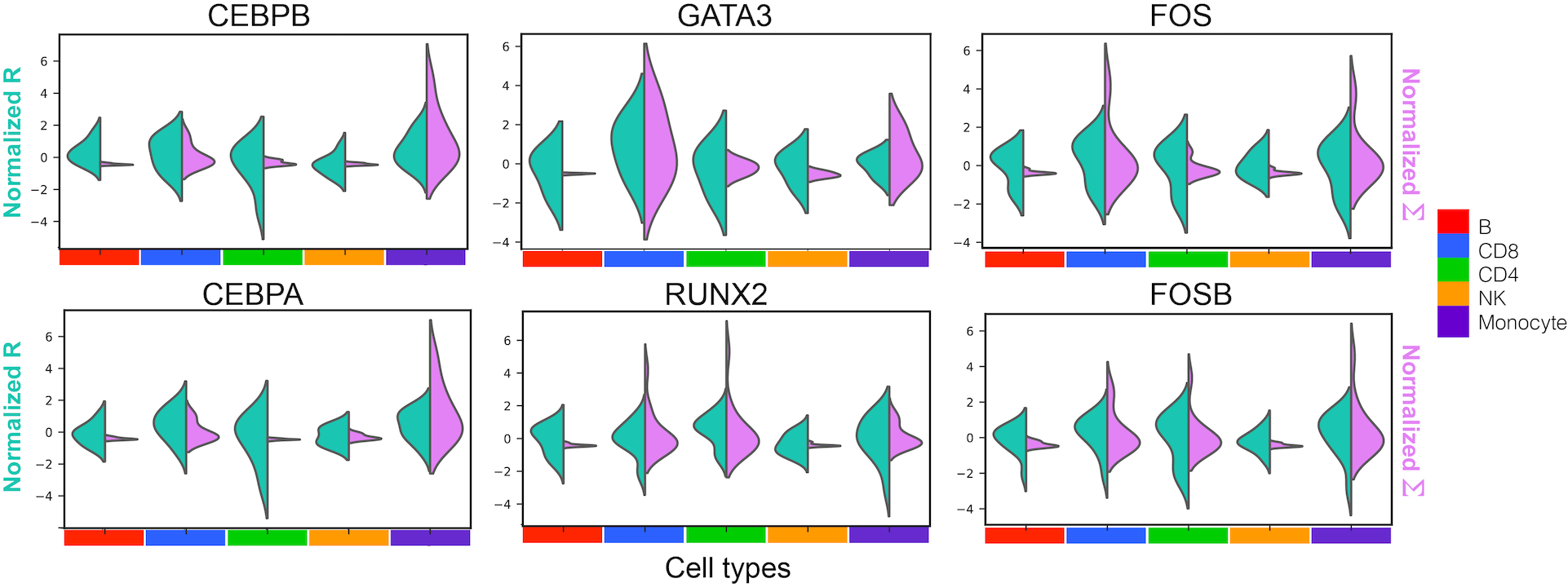} 
\caption[]{Identification of cell type-specific regulators. Violinplots of inferred cell type-specific regulatory function $R$ (green) for TFs that show variability across cell types, compared to inferred covariance $\Sigma$ (pink) between example TFs (regulators) and all of their target genes; positive and negative signs in $R$ denote activation and repression of expression, respectively; values have been z-score normalized to display both variables on the same y-axis.}
\label{fig:Rksbox}
\end{figure*}



\section{Inference} 
We applied the co-ordinate ascent mean field variational inference (CAVI) \citep{blei2017variational, ghahramani2001propagation} which assigns independent factors to the latent variables. The blueprint for corresponding CAVI updates are below and full derivations are presented in Supplementary section C.


\textbf{Variational E-step}  \\ \\
\textbf{a.}  $q^{*}( z_j) = \prod_k r_{jk}^{z_{jk}}$ where 
\begin{equation}
\begin{split}
 r_{jk} =  \mathbb{E}_{\pmb z}  {z_{jk}}  \propto  & \widetilde{|\beta_j \Sigma_k|^{-1}} \exp (-S_2)  \widetilde{\pi_k}, \  \sum_k r_{nk} =1 \\
  S_2 =& \tfrac{1}{2}\Big(\textit{tr}(\Sigma_k^{-1}\beta_j^{-1}\Sigma^{\prime \prime -1}) + \\
 & (\pmb \mu^{\prime \prime} - \alpha_j\pmb \mu_k)^T  (\beta_j\Sigma_k)^{-1} (\pmb \mu^{\prime \prime} - \alpha_j\pmb \mu_k) \Big) \\
%
%
 \widetilde{|\beta_j \Sigma_k|^{-1}}  := & \frac{1}{2}(-d \mathbb{E}_{\beta_j}\ln \beta_j + \\
 & d \ln (2) + \ln |R^*_k| + \sum_{i=1}^{d}\psi\Big(\frac{\gamma +1-i}{2}\Big))\\
\end{split}
\end{equation}\vspace{-5mm}

where $\ln \widetilde{\pi_k}  :=  \psi(\varphi_0) - \psi (\sum_k \varphi_k)$ and $\psi$ is the \textit{digamma} function and $R^*_k= (R_k + R_k^T)^2$.

\textbf{Variational M-step}  \vspace{-5mm}

\begin{equation}
\begin{split}
\textbf{b. } q^{*}(\pi_k) =& \pi_k \sim \text{Stick-breaking Beta}(1,\varphi) \\
\textbf{c. } q^{*}(\pmb \mu_k) =&  \exp \Big( -\tfrac{1}{2} \sum_j  r_{jk}  \Big(\textit{tr}(\Sigma_k^{-1}(\tfrac{\beta_j}{\alpha_j^2})^{-1}\Sigma^{\prime \prime -1}) \\
& +( \bar{\pmb {\mu_k}} - \frac{x_j}{\alpha_j})^T {(\frac{\beta_j}{\alpha_j^2} \Sigma_k)}^{-1}( \bar{\pmb {\mu_k}} - \frac{x_j}{\alpha_j}) \\
& + (\pmb\mu_k - \pmb{\mu}^\prime)^T \Sigma^{\prime -1}(\pmb \mu_k - \pmb{\mu}^\prime) \Big) + c \Big) \\
\end{split}
\end{equation} 
\begin{equation}
\begin{split}
\textbf{d. }  q^{*}(\Sigma_k^{-1}) =& \exp \Big( -\tfrac{1}{2}\sum_j r_{jk}\Big(\ln \beta_j^d + \ln |\Sigma_k| \\
& + \tfrac{1}{\beta_j} \Big(\textit{tr}(\Sigma_k^{-1}\Sigma^{\prime \prime -1}) \\
& + (\pmb{\mu}^{\prime \prime} - \alpha_j \pmb\mu_k)^T  (\Sigma_k)^{-1} (\pmb{\mu}^{\prime \prime} - \alpha_j\pmb \mu_k)  \Big) \Big) \\
&  + \ln |\Sigma_k^{-1}|^{\frac{\lambda-d-1}{2}} + \{-\frac{tr(R^{*-1}_k\Sigma_k^{-1})}{2}\} \\
& - \ln(2^{\lambda d/2} |R^{*-1}_k|^{-\lambda/2} \Gamma_d(\lambda/2)) +c \Big) \\
 \textbf{e.} q^{*}(\alpha_j) =&  \exp \Big(-\sum_k  r_{jk} S_2 \\
 +  \frac{1}{ \sqrt{2\delta^2\pi}} & \exp \Big(- \frac{(\ln \alpha_j -\nu)\mathbb{I}_{1\times 1}(\ln \alpha_j-\nu)}{2\delta^2} \Big) + c \Big) \\
\textbf{f. } q^{*}(\beta_j) =& \exp\Big(-\sum_k  r_{jk} S_2 \\ 
 + \frac{1}{\sqrt{2\theta^2\pi}} & \exp  \Big(- \frac{(\ln \beta_j 
 -\omega)\mathbb{I}_{1\times 1}(\ln \beta_j - \omega)}{2\theta^2} \Big)+ c \Big) \\
 \textbf{g. } q^*(\pmb \mu^\prime) \sim & \mathcal{N}  (\pmb \mu_{\mu^\prime}  ,\Sigma_{\mu^\prime})   \\
 \end{split}
\end{equation} 
\begin{equation}
\begin{split}
 \textbf{h. } q^*(\Sigma^{-1 \prime}) & \sim   \mathcal{W}(V_{\Sigma^{\prime-1}},d_{\Sigma^{\prime-1}})  \\
\textbf{i. }  q^*(R_k) = & \exp \Big( \{-\frac{tr(R^{*-1}_k\Sigma_k^{-1})}{2}\} \\  - \ln(2^{\lambda d/2} & |R^{*-1}_k|^{-\lambda/2} \Gamma_d(\lambda/2))  \\
  - \tfrac{1}{2\delta^2} \sum_i \sum_{i^\prime} & \Big( (R_k^{i,i^\prime}- S^{i,i^\prime}  M^{i,i^\prime} \pmb p_k^{g(i,i^\prime)})\mathbb{I}_{1\times 1}(R_k^{i,i^\prime}- \\
  & S^{i,i^\prime}  M^{i,i^\prime} \pmb p_k^{g(i,i^\prime)}) \Big)+ c\Big) \\ 
 \textbf{j.} q^*(\pmb p_k) = \exp & \Big(   (\bar c_t - \sum_k(\pi_k \pmb p_k))^T  (\zeta\mathbb{I})^{-1} (\bar c_t - \sum_k(\pi_k \pmb p_k))] \\
  + \ln & \Big(\tfrac{1}{\sqrt{2\pi}\Lambda}\exp(-\tfrac{1}{2} (\tfrac{\pmb p_k-\eta}{\Lambda})^2))\Big)   \\
  -\tfrac{1}{2\delta^2} & \sum_i \sum_{i^\prime}  \Big( (R_k^{i,i^\prime}  - S^{i,i^\prime}  M^{i,i^\prime} \pmb p_k^{g(i,i^\prime)})\mathbb{I}_{1\times 1} \\ 
 & (R_k^{i,i^\prime} - 
   S^{i,i^\prime}  M^{i,i^\prime} \pmb p_k^{g(i,i^\prime)}) \Big) + c \Big)
\end{split}
\end{equation} 
where $c$ is the integration constant (details are included in Supplementary section C). 
Since the E-step takes $\approx \mathrm{O}(d^3)$ due to three matrix inversions, we substitute this with $z_{\small{MAP}}(\pmb x) = \arg\max_z p(\pmb x|\pmb z)p(\pmb z|\pmb \pi)$ which takes $ \approx \mathrm{O}(d^2)$ when $\Sigma_k^{-1}$s and $\Sigma^{\prime \prime}$ are apriori Cholesky decomposed. 

Given the complexity of the model (refer to Supplementary section C), we implemented \emph{Symphony} using the probabilistic programming language Stan  \citep{carpenter2016stan}. Furthermore, for a scalable implementation applicable to real genomic data containing thousands of cells, we used the probabilistic programming language, Edward \citep{tran2016edward,tran2017deep} for variational EM (details and approximations presented in Supplementary section E). 





\section{Results}
%
%
%
\subsection{Synthetic Data} 
We first evaluated the performance in deconvolving epigenetic data, clustering cells, and inferring GRNs using data simulated from the \emph{Symphony} model. We simulated data for $n=100$ cells in $K=3$ clusters with $d$ ranging from $5$ to $20$ genes and $l=50$ using the \emph{Symphony} model. 
%

{\bf Inference of $\pmb{p}_k, R_k$.}
Figure \ref{fig:pscatter}(a) shows scatterplots of deconvolved peak heights ($\pmb p_k$) compared to actual data. 
We compared the performance of \emph{Symphony} to two other deconvolution methods: \emph{deconf}  \citep{repsilber2010biomarker}  which uses NMF,  and \emph{Dsection}  \citep{erkkila2010probabilistic} which is based on a Bayesian model.
While Dsection captures only the largest cluster, deconf underestimates the cluster-specific peak heights. The behavior of Dsection was reproducible across simulations, and is likely due to the lack of identifiability in the model for epigenetic data alone, as discussed above.

Figure \ref{fig:pboxplot} summarizes the error in estimating $\pmb p_k$s across  $10$ synthetic datasets with the same size as above.
This shows the value of incorporating expression data (view 1) in deconvolution of epigenetic data (view 2).
Figure \ref{fig:pboxplot} also shows a heatmap of inferred $R_k$ in one of the synthetic experiments as an example, compared to the actual $R_k$, confirming the ability of \emph{Symphony} in inferring GRNs.

{\bf Clustering performance.}
We then show the performance in clustering 
with integrating both views as compared to only using gene expression data (view 1) by computing F-scores across $10$ experiments with the same size as above. We compared the performance to BISCUIT \citep{prabhakaran2016dirichlet}, as well as other methods commonly used for clustering cells in single-cell gene expression data including DBscan \citep{satija2015spatial}, Phenograph \citep{levine2015data}, Spectral clustering  \citep{ng2002spectral} and k-means (with $K=3$) (Figure \ref{fig:pboxplot}). These results show  improvement over BISCUIT due the epigenetic extension of the model and significant improvement over other methods.
DBscan was unable to cluster the majority of cells, likely due to the small dimensionality of the feature space used in simulations.
This shows the value of incorporating epigenetic data (view 2) in improving clustering performance, as compared to using expression data (view 1) alone.
This has further value in biological interpretation of clusters as cell types that have both similar expression and similar underlying mechanisms driving expression.

\subsection{Genomic Data}
%
 We also evaluated the performance of \emph{Symphony} on real genomic data. We used previously published single-cell expression data for  peripheral blood mononuclear cells (PBMCs) from \citet{zheng2017massively} combined with ATAC-seq data for PBMCs from \citet{corces2016lineage}. 
 %
%
For single-cell expression data, we chose a subset of PBMCs from \cite{zheng2017massively} as $X$ containing $n=6825$ cells which  express  known gene markers for either monocytes, B cells or T cells and NK cells. We chose to focus on $d=28$ transcription factors which showed high standard deviation in expression data and are known to be lineage-defining factors.

For epigenetic data matching the above cell types, we generated $r=1$ mixture of epigenetic measurements with $l=1053$ peaks from real ATAC-seq data collected from the above sorted cell types in \citet{corces2016lineage}, with weights $\pi_k$ proportional to frequency of cell types in blood, and used this as  observed epigenetic data $C$. We fixed the clustering in this experiment using Phenograph-derived assignments which we mapped onto nearest cell types to match with epigenetic data \cite{levine2015data}. 

We determined non-zero entries in $M$ from ATAC-seq using the FIMO algorithm \citep{grant2011fimo}, which scans the sequence under the ATAC-seq peak for the occurrence of a motif. We associated a peak with the target gene closest in genomic distance to the peak in these experiments. This assignment is independent of the model structure and can be manually defined by the user. 

In the following tests, we used a scalable implementation with Edward \cite{tran2016edward}  detailed in Supplementary section E. Figure \ref{fig:peakdeconv_edward} shows the performance of this implementation on a larger number of cells and genes.

{\bf Cell type characterization.}
%
In this test, we pre-imputed and normalized data, and fix BISCUIT-derived normalization parameters ($\alpha,\beta$). Specifically, we normalized and imputed $\pmb x_j$ for each cell $j$ by transforming it to $\pmb y_j$ with 
$\pmb y_j = A \pmb x_j + b$, and setting $A=I/\beta_j$ and $b=(I-\alpha_j A)\pmb \mu_k$ with BISCUIT -inferred parameters. This transformation corrects for cell-specific technical effects captured by $\alpha_j.\beta_j$, as $\pmb y_j \sim \mathcal{N}(\pmb \mu_k, \Sigma_k)$ while $\pmb x_j \sim \mathcal{N}(\alpha_j\pmb \mu_k, \beta_j\Sigma_k)$ (Figure \ref{fig:Plate_model}) \citep{prabhakaran2016dirichlet}.
Figure \ref{fig:pbmc} (a) shows t-SNE projections \citep{maaten2008visualizing} of $\pmb y_j$s after normalizing expression data based on inferred parameters $\pmb \mu_k, \Sigma_k, \pmb \alpha, \pmb \beta$ where cells are colored by cluster assignments $z_j$s for $K=5$ distinct cell populations. 
Figure \ref{fig:pbmc} (b) shows normalized expression of known marker genes used to characterize the clusters as monocyte, B, T and NK cell types.

{\bf Deconvolving epigenetic data.}
Figure \ref{fig:pbmc} (c,d) show inferred peak heights $\pmb p_k$ for all clusters using \emph{Symphony} compared to ground truth cell type specific peak heights (measured with ATAC-seq from sorted cell types). 
Supplementary Figure \ref{fig:tracks} shows an example genomic region with differential peaks  for three of these cell types showing distinct epigenetic profiles.
Projection of peak heights to principal components of ground truth peaks (excluding peaks un-constrained by expression data and peaks which show 0 accessibility in some cell types, to show performance at deciphering magnitudes) shows superior performance in deconvolving all subsets of cells except for the smallest population (NK cells, $<5$\% of cells).
 The small error between estimated $\pmb p_k$s and ground truth cell type-specific peak heights confirms that our model is a good fit for the biological mechanism of regulation. We also evaluated the deconvolution of epigenetic data using \emph{deconf} (Figure \ref{fig:pbmc} (c)) which shows under-estimation or inaccurate estimation of peak heights.
We could not test \emph{Dsection} as the number of clusters exceeds the number of replicates. 


{\bf Inferring GRNs.}
The main advantage of \emph{Symphony} is the inference of GRNs. 
Figure \ref{fig:fitprior} shows \emph{Symphony} can successfully learn cell type-specific covariances comparable to empirical covariances. The gene-gene covariance matrix is then  explained by direct regulation as well as the propagated impact of regulation through $(R+R^T)^2$, which we previously validated is capable of capturing epigenetic information through the accuracy of its prior $p$.    Furthermore, Figure \ref{fig:sortedR} reveals that inferred regulatory functions are explained by either TF binding strength (peak height) or TF-gene covariance or both. 
Figure \ref{fig:networks} shows the inferred GRNs between TFs with strongest inferred links ($|R_k|>4.5$) in each cluster. The differences in the structure of the networks suggests different mechanisms driving cell type-specific expression. 
%

%

We observe numerous regulatory interactions that are variable across clusters.
Figure \ref{fig:Rksbox} shows examples of TF-gene interactions ($R$) that are also supported by known literature. It can be seen that regulatory functions are partially supported by gene-gene covariances ($\Sigma$). 
We observe $CEBPA,CEBPB$ differentially regulating target genes in monocytes (cluster 5). A recent study \cite{jaitin2016dissecting} has shown that knock-outs of $CEBPB$ block monocyte differentiation.
%

We observe regulatory edges in the GRN for T cells between Transcription Factors (TFs) $GATA3,RUNX2$ and their target genes, with minimal interaction in B cells and NK cells (Figure \ref{fig:Rksbox}), and indeed these TFs are known to be associated with activation of cytotoxic T cells \cite{pearce2003control} and CD8 T cell development \cite{woolf2003runx3}.
We also observe cases such as $FOS, FOSB$ with different regulatory functions despite belonging to the same TF family, showing an example of how expression-derived information can further distinguish genetic interactions which cannot be immediately deciphered from epigenetic data.
%
%


\section{Conclusion}
We present a hierarchical Bayesian mixture model named \emph{Symphony} that infers clusters of cells representative of  cell types and gene regulatory networks (GRNs) specific to cell types. 
This is done  by modeling the regulatory mechanism  driving gene expression in each cell type, and assuming two observations as two views from the system: epigenetic measurements, which are informative of network edges and single-cell gene expression data, informative of network node activity.
To the best of our knowledge, this is the first computational method that integrates single-cell expression data with epigenetic data. We provide theoretical  justifications for the model and an EM-VI procedure.
\emph{Symphony} shows great performance in clustering cells, deconvolving epigenetic profiles and inferring GRNs in both synthetic and real data from peripheral blood cells and shows superiority to other methods that only address one of these problems. Further, \emph{Symphony} was able to deconvolve epigenetic data when only one replicate was available through integration of expression data, a potentially common task which would be challenging for any source separation technique. Future iterations of the experiments will allow a \emph{Symphony} -derived clustering of PBMCs to improve the mapping of cells to cell types, particularly for more similar cell types such as CD4+ and CD8+ T cells and when few genes are considered.
Applied to the growing single-cell datasets, Symphony can reveal cell type-specific regulation in normal cells as well as disrupted regulation in cancerous cells. 





\bibliography{refs}
\bibliographystyle{icml2019}

\appendix




\onecolumn

\section*{Supplementary Materials for A Nonparametric Multi-view Model for Estimating Cell Type-Specific Gene Regulatory Networks}

\section{Supplementary Figures}


\begin{figure}[bth]
\centering
\includegraphics[width=.85\linewidth]{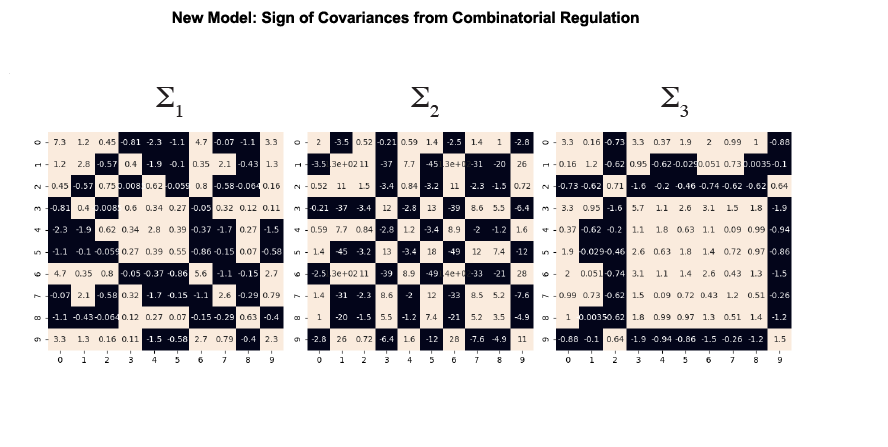}
\caption[]{Examples of synthetic $\Sigma_k$ simulated from the same $S$ matrix for clusters $k=1,2,3$, showing variability in sign of $\Sigma_k$ that can capture the impact of combinatorial regulations.} 
\label{fig:covsim}
\end{figure}

\begin{figure}[bth]
\centering
\includegraphics[width=.85\linewidth]{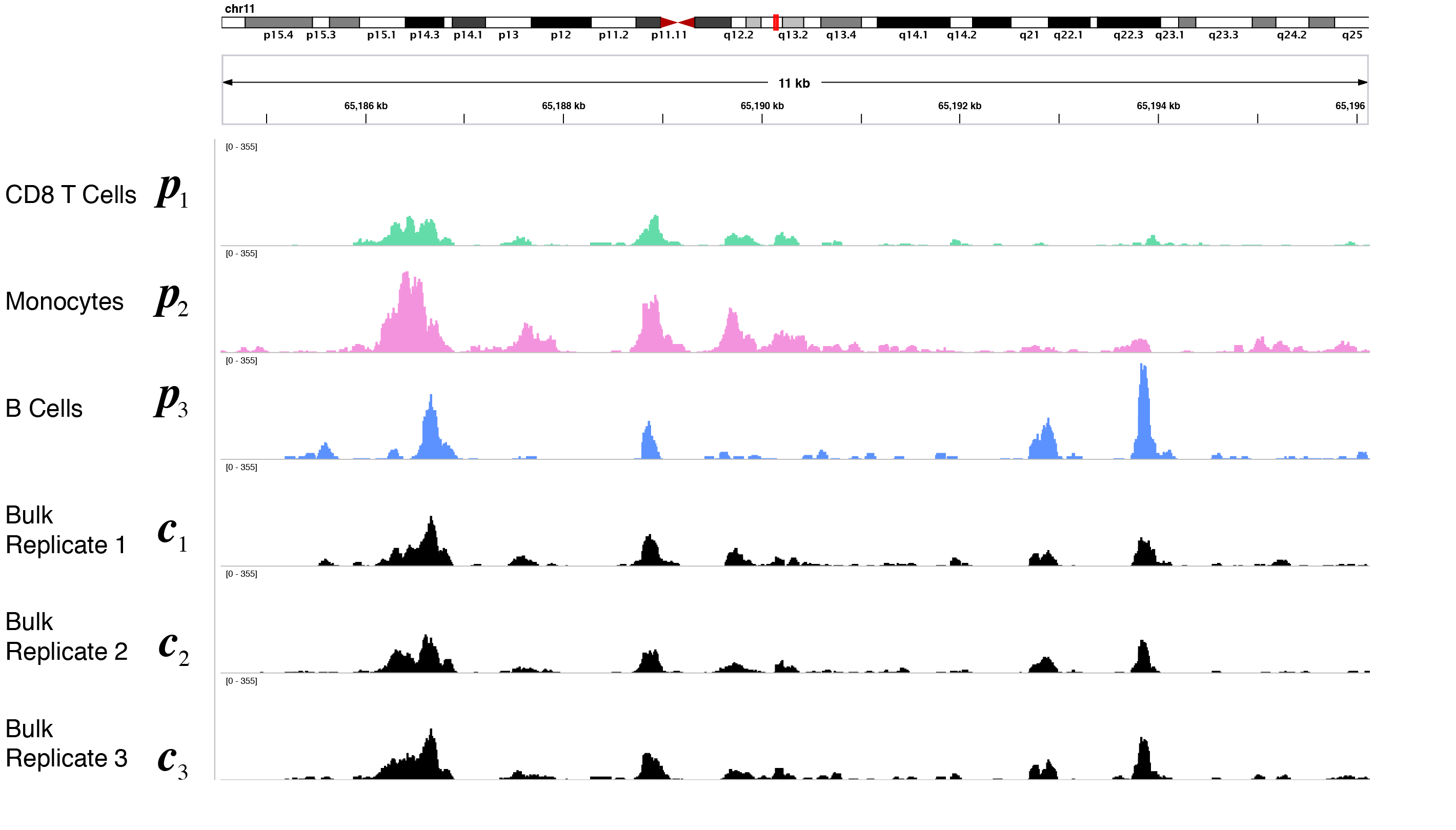}
\caption[]{Epigenetic (ATAC-seq) data visualized for the  three cell types used in section 7 showing an example region with differential peak heights $\pmb p_k$s across the cell types and examples of simulated bulk data $\pmb c_t$ for $t=1,2,3$ from the weighted sum.} 
\label{fig:tracks}
\end{figure}

\begin{figure*}[tbh]
\centering
\includegraphics[width=1\linewidth]{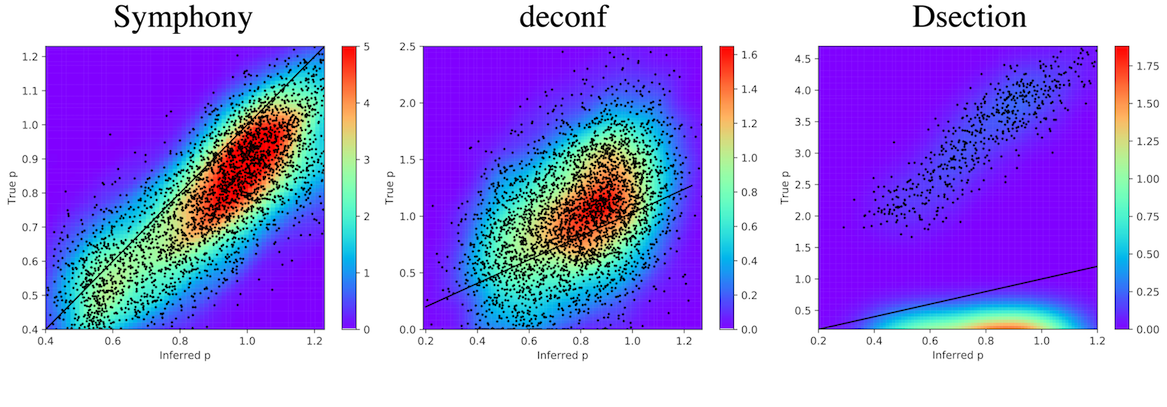}\vspace{-5mm}
\caption[]{Performance in deconvolving epigenetic data with Edward implementation. Estimated peak heights $\pmb p_k$ using \emph{Symphony} for $n=4000$ simulated cells in $K=3$ clusters with $d=100$ genes and $l=550$ versus true peak heights, compared to two other deconvolution methods: deconf \citep{repsilber2010biomarker} and Dsection \citep{erkkila2010probabilistic}; each dot represents a genomic region; heatmap shows density. } 
\label{fig:peakdeconv_edward}
\end{figure*}


\section{Extended Theory}


\textbf{Lemma 1} \textit{The epigenetic model $f(\pmb c|\pmb{p}_k, \pi_k)$ is non-identifiable} 
 \label{lemdeconv} 
 \\
\textit{Proof sketch.} Due to having $Kl+K$ unknowns in the mean parameters while having an $l$ dimensional Normal distribution, where $K$ is the maximum number of allowed clusters, we have an under-determined problem. Thus, we can provide multiple parameter sets $\pi_k$ and $\pmb{p}_k$ leading to the same Normal distribution  for $\pmb c$. 


\textbf{Lemma 3}
\textit{Square of a symmetric matrix $H$ gives a symmetric positive semi-definite matrix $L$.}
\label{lemsympsd}
\\
\textit{Proof. } We show that there exists a symmetric positive semi-definite matrix $L$ \emph{iff} there exists a symmetrix matrix $H$ that satisfies $H^2=L$. $H,L \in \mathrm{R^{d\times d}}$.
Orthogonal diagonalization of $L$ gives $QLQ^{-1} = D^2$ where $Q$ is the orthogonal matrix and $D$ is the square root diagonal matrix $diag(l_1^{\frac{1}{2}},\cdots, l_d^{\frac{1}{2}})$. If there exists a matrix $H$ where $H= Q * diag(l_1^{\frac{1}{2}},\cdots, l_d^{\frac{1}{2}}) * Q^{-1} = QDQ^{-1}$, then $QHQ^{-1} = D$. We now write: 
\begin{equation*}
\begin{split}
QLQ^{-1} &= D^2\\
&=diag(l_1^{\frac{1}{2}},\cdots, l_d^{\frac{1}{2}}) * diag(l_1^{\frac{1}{2}},\cdots, l_d^{\frac{1}{2}})\\
&= QHQ^{-1}QHQ^{-1} =QH^2Q^{-1}\\
\end{split}
\end{equation*}
showing $H^2=L$. Next assume $H$ is symmetric and therefore all its eigenvalues are real. For some eigenvalue $h$ of $H$, $l$ is an eigenvalue of $L$ \emph{iff} $l=h^2$ implying all eigenvalues of $L \in \mathrm{R}^{*} $ where $\mathrm{R}^* = \{0\} \cup \mathrm{R}^+$. Further, $L$ is symmetric given $H$ is symmetric. We now have $L$ as symmetric and has non-negative eigenvalues proving that $L$ is positive semi-definite.    
$\blacksquare$


%

%
%

\textbf{Lemma 4}
\textit{In the reduced model: $f(X^0,C|\pmb \beta, \pmb \mu_k^0, \Sigma_k^0, \pmb \alpha, \pmb z, \pmb p_k, R_k^0, \pi_k, \zeta)$, $\beta$s are identifiable under the conditions of: $\forall j: \pmb\mu_k \ge \pmb\mu^\prime + diag(\Sigma^\prime)(\alpha_j-\nu)/\delta$ without the need for condition on $\beta$s.}
\label{lembeta}

\textit{Proof sketch. } Using Lemma 2 and focusing on the marginal distribution $f(X^0|\pmb \beta, \pmb \mu_k^0, \Sigma_k^0, \pmb \alpha, \pmb z, \pi_k)$ we know that $\beta_j \Sigma_k^0$ and $\pi_k$ are identified for all $j$ and $k$.
Hence, $\beta_j^{1/2}\Sigma \pi_k (\Sigma_k^0)^{1/2}$ is identified. Using Lemma 3, $(R_k + R_k^T)^2$ is non-negative semi definite which allows us to define $(\Sigma_k^0)^{1/2}$ from the Wishart distribution. 

Therefore, 
$\beta_j^{1/2}\Sigma \pi_k (R_k + R_k^T)$ is identifiable.
We also know that $\Sigma \pi_k \pmb p_k$ is identified through the marginal distribution for $C$. 
Thus, putting the above together, it is not possible to have multiple values for $\beta_j$ since that would require the sum $\Sigma \pi_k (R_k + R_k^T)$ to be different for two sets of parameters while each element of this sum can be written as elements of $\Sigma \pi_k \pmb p_k$ which is identified. $\blacksquare$





\textbf{Lemma 5}
\textit{For a given $\pmb \beta = \beta^*$, identifiability of: $f(X,C|\pmb \mu_k, \Sigma_k, \pmb \alpha, \pmb \beta=\pmb \beta^*, \pmb z, \pmb p_k, R_k, \pi_k, \zeta)$ is guaranteed if $ \forall j: \pmb\mu_k \ge \pmb\mu^\prime + diag(\Sigma^\prime)(\alpha_j-\nu)/\delta$.} 

\textit{Proof sketch. } Parameters related to BISCUIT are identifiable using Lemma  2
without any of conditions on $\beta$s used in BISCUIT's proof, since $\beta$s are all given. 

It remains to show identifiability for parameters $\pmb p_k, R_k, \zeta$ from integrated model on $C$. $\zeta$ is identified due to Normality of $C$. Identifiability of $\pmb p_k$ will lead to identifiability of $R_k$. We focus on proving identifiability of $\pmb p_k$. 

Using the identifiability of $\Sigma_k$ we have $d(d-1)/2$ equations based on generation of $\Sigma_k$ based on $\pmb p_k$. Considering that $\pmb p_k$ has $l$ unknowns and the relationships are at most polynomials of degree 2, as long as $d(d-1)/2 > l^2$ we have an overdetermined system of equations to identify $\pmb p_k$ from $\Sigma_k$.
$\blacksquare$

%

\textbf{Theorem 6}
\textit{The full model $f(X,C|\Theta)$ where $\Theta := \{ \pmb \mu_k, \Sigma_k, \pmb \alpha, \pmb \beta, \pmb z, \pmb p_k, R_k, \pi_k, \zeta\}$ is identifiable if $ \ \forall j: \pmb\mu_k \ge \pmb\mu^\prime + diag(\Sigma^\prime)(\alpha_j-\nu)/\delta$.}
\\
\textit{Proof sketch.} To prove identifiability of $\beta$, we use Lemma 5 on the following reduced distribution of $f(X,C|\Theta)$: $f(X^0,C|\pmb \beta, \pmb \mu_k^0, \Sigma_k^0, \pmb \alpha, \pmb z, \pmb p_k, R_k^0, \pi_k, \zeta)$.
Given the identified $\beta$s from Lemma 5, we use Lemma 6 to conclude identifiability of the rest of the parameters of the full model $f(X,C|\Theta)$, as desired. $\blacksquare$

\section{Variational Inference update equation derivations}

\subsection{Joint distribution for X and C}

We use the graphical model in Figure \ref{fig:Plate_model} to write down the variational inference equations for Symphony. Note that this is constructed based on conditionally-conjugate priors for $\pmb \mu$ and $\Sigma$ and on space discretised by $\pmb{z}=\{z_1, \cdots, z_n\}$ where $z_j =\{j\}, j=[1, \cdots, k]$. The joint is written based on the Markov blanket for each parameter.

\begin{equation}
\begin{split}
p(X, C, \pmb z, \pmb \pi,\pmb \mu, \Sigma,\pmb \mu^\prime, \Sigma^\prime, R, S, M , \lambda, \pmb \alpha, \pmb \beta ,\pmb{p},\eta,\gamma,\Lambda, \zeta) &= p(X|\pmb z,\pmb \mu, \Sigma, \pmb \alpha, \pmb \beta) p(C|\pmb p, \pmb \pi, \zeta) \\
& p(\pmb z|\pmb \pi)p(\pmb \pi|\pmb \varphi) \\
& p(\pmb \mu^\prime|\pmb \mu^{\prime\prime}, \Sigma^{\prime\prime})p(\Sigma^{-1 \prime}|\Sigma^{-1 \prime\prime}, d)\\
& p(\pmb \alpha| \nu, \delta)p(\pmb \beta|\omega, \theta)\\
& \prod_k p(\pmb \mu_k|\pmb \mu^\prime, \Sigma^\prime)p(\Sigma^{-1  }_k| (R_k+R_k^T)^2,\gamma)p( R_k|SM\pmb{p_k}, \lambda) p(\pmb p_k| \eta, \Lambda)\\
\end{split}
\label{eq: Joint}
\end{equation}

Next we expand each term in the RHS of Equation \ref{eq: Joint}. 

 \begin{equation}
\begin{split}
p(X|\pmb z,\pmb \mu, \Sigma,\pmb \alpha, \pmb \beta) &=\prod_j p(\pmb x_j | z_j,\pmb \mu, \Sigma, \alpha_j, \beta_j)\\
&=\prod_j \prod_k p(\pmb x_j | z_{jk},\pmb \mu_k,\Sigma_k, \alpha_j, \beta_j)^{z_{jk}}\\
&=\prod_j \prod_k \mathcal{N}(\pmb x_j|\alpha_j \pmb \mu_k, \beta_j \Sigma_k)^{z_{jk}}\\
\end{split}
\label{eq: RHS_X}
\end{equation}

\begin{equation}
\begin{split}
p(\pmb z|\pmb \pi) &=\prod_j p(z_j | \pmb \pi)\\
&=\prod_j \prod_k p(z_{jk} | \pi_k)^{z_{jk}}\\
&=\prod_j \prod_k \mathrm{Mult}(z_{jk} | \pi_k)^{z_{jk}}\\
&=\prod_j \prod_k  \pi_k^{z_{jk}}\\
\end{split}
\label{eq: RHS_Z}
\end{equation}

\begin{equation}
\begin{split}
p(\pmb \pi|\pmb \varphi) &=\mathrm{Dir} (\pmb \pi | \varphi_1, \cdots, \varphi_K) \\
&=\mathrm{Dir} (\pmb \pi | \varphi_0) \\
&=\frac{1}{B(\varphi_0)}\prod_k \pi_k^{\varphi_0-1}
\end{split}
\label{eq: RHS_pi}
\end{equation}
where $\sum_k \pi_k = 1 \quad \text{and} \quad \varphi_1=\cdots= \varphi_K $ for symmetric prior.
\textcolor{blue}.

In the experiments, we used:
\begin{equation}
\begin{split}
p(\pmb \pi| \varphi) &=\mathrm{Stick} (\pmb \pi | \varphi) \\
p( \pi_k^\prime| \varphi)&=\mathrm{Beta} (\pi_k^\prime | 1, \varphi) \\
 \pi_k &= \pi_k^\prime \prod_i^{K-1}(1-\pi_i^\prime) \\
\end{split}
\label{eq: RHS_Stick_pi}
\end{equation}
where $\sum_k \pi_k = 1 $ and $\pi_k$ is the length of the \textit{k-th} stick / proportion of the \textit{k-th} cluster in stick breaking.

\begin{equation}
\begin{split}
p(C | \pmb p,\pmb \pi, \zeta) &= \prod_t^r p(c_t | \pmb p,\pmb \pi, \zeta)\\ 
&= \prod_t^r \sum_k p(c_t | \pmb p,\pmb \pi, \zeta)\\
&= \prod_t^r \sum_k \mathcal{N}(c_t | \pi_k p_k, \zeta \mathrm{I})\\
\end{split}
\label{eq: RHS_C}
\end{equation}

\begin{equation}
\begin{split}
p(\pmb \mu^\prime | \pmb \mu^{\prime\prime},\Sigma^{\prime\prime}) &=\mathcal{N}(\pmb \mu^\prime | \pmb \mu^{\prime\prime},\Sigma^{\prime\prime})\\
\end{split}
\label{eq: RHS_muprime}
\end{equation}

\begin{equation}
\begin{split}
p( \Sigma^{\prime -1} | \Sigma^{\prime \prime -1},d) &=\mathcal{W}(  \Sigma^{\prime -1} | \Sigma^{\prime \prime -1},d)\\
\end{split}
\label{eq: RHS_Sigmaprime}
\end{equation}

\begin{equation}
\begin{split}
p(\pmb \alpha|\nu,\delta) &=\prod_j p( \alpha_j|\nu,\delta)\\
&=\prod_j \mathrm{logNormal}( \alpha_j|\nu,\delta)\\
\end{split}
\label{eq: RHS_alpha}
\end{equation}

\begin{equation}
\begin{split}
p(\pmb \beta | \omega,\theta) &=\prod_j p( \beta_j|\omega,\theta)\\
&=\prod_j \mathrm{logNormal}( \beta_j|\omega,\theta)\\
\end{split}
\label{eq: RHS_beta}
\end{equation}

\begin{equation}
    \begin{split}
        \prod_k p(\pmb \mu_k|\pmb \mu^\prime, \Sigma^\prime)p( \Sigma^{-1  }_k| R_k,\gamma)p( R_k| S  M\pmb{p}_k, \lambda) p(\pmb p_k| \eta, \Lambda) &= \prod_k \mathcal{N}(\pmb \mu_k | \pmb \mu^\prime,\Sigma^\prime) \mathcal{W}(  \Sigma^{-1}_k|(R_k+R_k^T)^2,\gamma)\\\prod_i \prod_{i^\prime} \mathcal{N}( R_k^{i,i^\prime}| S^{i,i^\prime}  M^{i,i^\prime} \pmb p_k^{g(i,i^\prime)},\lambda) \mathrm{trunc}\mathcal{N}(\pmb{p_k}|\eta, \Lambda, \pmb 0, +\infty) \\
    \end{split}
\end{equation}

\subsection{Variational distributions}

We now write the factorized distribution $q$ which will approximate the joint distribution in Equation \ref{eq: Joint}. We follow Chapter 10 of Bishop \cite{bishop2006pattern} and the Matrix cookbook Section 8.2 \cite{petersen2008matrix}.

\begin{equation}
\begin{split}
q(\pmb z, \pmb \pi,\pmb \mu, \Sigma,\pmb \mu^\prime, \Sigma^\prime, R, S, M , \lambda, \pmb \alpha, \pmb \beta ,\pmb{p},\eta,\gamma,\Lambda, \zeta | X,C) &= \underbrace{q(\pmb z | X,C)}_{Variational \,\, E-step} \\ & \quad \underbrace{q( \pmb \pi,\pmb \mu,  \Sigma,\pmb \mu^\prime, \Sigma^\prime, R, S, M , \lambda, \pmb \alpha, \pmb \beta ,\pmb{p},\eta,\gamma,\Lambda, \zeta |X,C)}_{Variational \,\, M-step} \\
\end{split}
\label{eq: Joint_factored}
\end{equation}

The sequential update equations can be written in terms of the E-step and M-step as follows:

\paragraph{Variational E-step.}
Take the expectation of the log of the joint distribution with respect to $\pmb \Theta := \{\pmb \pi,\pmb \mu, \Sigma,\pmb \mu^\prime, \Sigma^\prime, R, S, M , \lambda, \pmb \alpha, \pmb \beta ,\pmb{p},\eta,\gamma,\Lambda, \zeta\}$ i.e. all other parameters except $\pmb z$.
We have:

\begin{equation}
\begin{split}
\ln q^*(\pmb z | X, C) &= \mathbb{E}_{\pmb \Theta}[\ln p(X,C,\pmb \pi,\pmb \mu, \Sigma,\pmb \mu^\prime, \Sigma^\prime, R, S, M , \lambda, \pmb \alpha, \pmb \beta ,\pmb{p},\eta,\gamma,\Lambda ,\zeta \ )] + Const \\
&= \mathbb{E}_{\pmb \Theta}[ \ln p(X|\pmb z,\pmb \mu, \Sigma,\pmb \alpha, \pmb \beta) + p(C|\pmb p, \pmb \pi, \zeta) + \ln p(\pmb z|\pmb \pi)+ \ln p(\pmb \pi|\pmb \varphi) +\\
&  \ln p(\pmb \mu^\prime|\pmb \mu^{\prime\prime}, \Sigma^{\prime\prime}) + \ln p(\Sigma^{-1\prime}|\Sigma^{-1\prime\prime}, d) +\\
&  \ln p(\pmb \alpha| \nu, \delta) + \ln p(\pmb \beta|\omega, \theta)] + \\
& \sum_k \Big( \ln p(\pmb \mu_k|\pmb \mu^\prime, \Sigma^\prime)+ \ln p(\Sigma^{-1  }_k| (R_k+R_k^T)^2,\gamma)+ \ln p( R_k|SM\pmb{p_k}, \lambda)+ \ln  p(\pmb p_k| \eta, \Lambda) \Big)] + Const
\end{split}
\label{eq: Estep_1}
\end{equation}

Taking terms in $\pmb z$ alone:
\begin{equation}
\begin{split}
\ln q^*(\pmb z | X,C) &= \mathbb{E}_{\pmb \Theta}[ \ln p(X|\pmb z,\pmb \mu, \Sigma,\pmb \alpha, \pmb \beta) + \ln p(\pmb z|\pmb \pi)] + Const\\
&=\mathbb{E}_{\pmb \Theta}[\ln \prod_j \prod_k \mathcal{N}(\pmb x_j|\alpha_j \pmb \mu_k, \beta_j \Sigma_k)^{z_{jk}} + \ln \prod_j \prod_k  \pi_k^{z_{jk}}] + Const\\
&=\mathbb{E}_{\pmb \Theta}[ \sum_j \sum_k z_{jk} \ln \mathcal{N}(\pmb x_j|\alpha_j \pmb \mu_k, \beta_j \Sigma_k) +  \sum_j \sum_k  z_{jk} \ln\pi_k] + Const\\
&=\mathbb{E}_{\pmb \Theta}[ \sum_j \sum_k z_{jk} [\ln \mathcal{N}(\pmb x_j|\alpha_j \pmb \mu_k, \beta_j \Sigma_k) + \ln\pi_k]] + Const\\
&=\mathbb{E}_{\pmb \Theta}[ \sum_j \sum_k z_{jk} [(-\frac{d}{2}\ln 2\pi + \frac{1}{2}\ln |\beta_j \Sigma_k|^{-1} - \frac{1}{2}(\pmb x_j - \alpha_j\pmb \mu_k)^T (\beta_j\Sigma_k)^{-1} (\pmb x_j - \alpha_j\pmb \mu_k)) + \ln\pi_k]] + Const\\
&= \sum_j \sum_k z_{jk} [\mathbb{E}_{\pmb \Theta}(-\frac{d}{2}\ln 2\pi) + \mathbb{E}_{\pmb \Theta}(\frac{1}{2}\ln |\beta_j \Sigma_k|^{-1}) \\
&- \mathbb{E}_{\pmb \Theta}(\frac{1}{2}(\pmb x_j - \alpha_j\pmb \mu_k)^T (\beta_j\Sigma_k)^{-1} (\pmb x_j - \alpha_j\pmb \mu_k)) + \mathbb{E}_{\pmb \Theta}(\ln\pi_k)] + Const\\
&= \sum_j \sum_k z_{jk} \ln \Delta_{jk} + Const\\
&\propto \sum_j \sum_k z_{jk} \ln \Delta_{jk}
\end{split}
\label{eq: Estep_2}
\end{equation}

Taking exponentials on both sides of Equation \ref{eq: Estep_2}:
\begin{equation}
\begin{split}
 q^*(\pmb z | X,C) &\propto \prod_j \prod_k  \Delta_{jk}^{z_{jk}}
\end{split}
\label{eq: Estep_3}
\end{equation}

where 
\begin{equation}
\ln \Delta_{jk} := -\frac{d}{2}\ln 2\pi + \underbrace{\mathbb{E}_{\pmb \Theta}(\frac{1}{2}\ln |\beta_j \Sigma_k|^{-1})}_{S1} - \underbrace{\mathbb{E}_{\pmb \Theta}(\frac{1}{2}(\pmb x_j - \alpha_j\pmb \mu_k)^T (\beta_j\Sigma_k)^{-1} (\pmb x_j - \alpha_j\pmb \mu_k))}_{S2} + \underbrace{\mathbb{E}_{\pmb \Theta}(\ln\pi_k)}_{S3}
\label{eq: Delta}
\end{equation}

Let us now expand the expectations given as S1, S2 and S3 in Equation \ref{eq: Delta}.

\begin{enumerate}
\item 
\begin{equation}
\begin{split}
S1&:=\mathbb{E}_{\pmb \Theta}(\frac{1}{2}\ln |\beta_j \Sigma_k|^{-1})\\
&=\frac{1}{2}\mathbb{E}_{(\beta_j, \Sigma_k)}(\ln |\beta_j \Sigma_k|^{-1})\\
&=\frac{1}{2}\mathbb{E}_{(\beta_j, \Sigma_k)}(\ln \beta_j^{-d} |\Sigma_k^{-1}|)\\
&=\frac{1}{2}\mathbb{E}_{(\beta_j, \Sigma_k)}(\ln \beta_j^{-d} +\ln|\Sigma_k^{-1}|)\\
&=\frac{1}{2}(\mathbb{E}_{(\beta_j, \Sigma_k)}\ln \beta_j^{-d} +\mathbb{E}_{(\beta_j, \Sigma_k)}\ln|\Sigma_k^{-1}|)\\
&=\frac{1}{2}(-d \mathbb{E}_{\beta_j}\ln \beta_j +\mathbb{E}_{\Sigma_k}\ln|\Sigma_k|)\\
&=\frac{1}{2}(-d \mathbb{E}_{\beta_j}\ln \beta_j + d \ln (2) + \ln |(\pmb R_k + \pmb R_k^t)^2| + \sum_{i=1}^{d}\psi\Big(\frac{\gamma_k +1-i}{2}\Big))\\
&\equiv \ln \widetilde{|\beta_j \Sigma_k|^{-1}}\\
\end{split}
\end{equation}
where $\psi(.)$ is the digamma function and equals $\tfrac{d}{dx} \log \Gamma(x)$ and $\gamma_k:=\gamma +n_k$. \textbf{Runtime complexity}$\sim \mathrm{O}(d^3)$

\item 

\begin{equation}
\begin{split}
S_2&:=\mathbb{E}_{\pmb \Theta}[\frac{1}{2}(\pmb x_j - \alpha_j\pmb \mu_k)^T (\beta_j\Sigma_k)^{-1} (\pmb x_j - \alpha_j\pmb \mu_k)]\\
&=\mathbb{E}_{(\Sigma_k,\beta_j,\alpha_j,\mu_k)}[\frac{1}{2}(\pmb x_j - \alpha_j\pmb \mu_k)^T (\beta_j\Sigma_k)^{-1} (\pmb x_j - \alpha_j\pmb \mu_k)]\\
&=\tfrac{1}{2}\sum_i\sum_j (\beta\Sigma_{k_{ij}})^{-1}\mathbb{E}[(\pmb x_j - \alpha_j\pmb \mu_k)^T  (\pmb x_j - \alpha_j\pmb \mu_k)] \quad \text{(linearity of expectation property)}\\
&=\tfrac{1}{2}\sum_i\sum_j (\beta\Sigma_{k_{ij}})^{-1} (\Sigma^{\prime \prime -1} + (\pmb \mu^{\prime \prime} - \alpha_j\pmb \mu_k)^T  (\pmb \mu^{\prime \prime} - \alpha_j\pmb \mu_k)) \quad \text{(via covariance formula)}\\
&=\tfrac{1}{2}\sum_i\sum_j (\beta\Sigma_{k_{ij}})^{-1} (\Sigma^{\prime \prime -1}) + \tfrac{1}{2}\sum_i\sum_j (\beta\Sigma_{k_{ij}})^{-1}(\pmb \mu^{\prime \prime} - \alpha_j\pmb \mu_k)^T  (\pmb \mu^{\prime \prime} - \alpha_j\pmb \mu_k) \quad \text{($\Sigma_k$ symmetricity)}\\
&=\tfrac{1}{2}\Big(\textit{trace}(\Sigma_k^{-1}\beta^{-1}\Sigma^{\prime \prime -1}) + (\pmb \mu^{\prime \prime} - \alpha_j\pmb \mu_k)^T  (\beta\Sigma_k)^{-1} (\pmb \mu^{\prime \prime} - \alpha_j\pmb \mu_k) \Big)\\
\end{split}
\end{equation}
\textbf{Runtime complexity}$\sim \mathrm{O}(d^3)$

\item 
\begin{equation}
\begin{split}
S3&:=\mathbb{E}_{\pmb \Theta}[\ln\pi_k]\\
&=\mathbb{E}_{\pmb \pi}[\ln\pi_k]\\
&=\psi(\varphi_0) - \psi (\sum_k \varphi_k)\\
&\equiv \ln \widetilde\pi_k
\end{split}
\label{eq: S3}
\end{equation}
where $\psi$ is the \textit{digamma} function and $\psi(t) = \frac{d}{d_0}\ln (\Gamma(t)) = \frac{\Gamma^\prime(t)}{\Gamma(t)}$. \textbf{Runtime complexity}$\sim \mathrm{O}(1)$
\end{enumerate}

Therefore, 
\begin{equation}
\ln \Delta_{jk} := -\frac{d}{2}\ln 2\pi + \ln \widetilde{|\beta_j \Sigma_k|^{-1}} - \mathbb{E}_{\pmb \Theta}[\frac{1}{2}(\pmb x_j - \alpha_j\pmb \mu_k)^T (\beta_j\Sigma_k)^{-1} (\pmb x_j - \alpha_j\pmb \mu_k)] + \ln \widetilde\pi_k
\label{eq: log_Delta}
\end{equation}

We require that $q^*(\pmb z | X)$ is normalised and that for every observation $j$, there is only one non-zero $z_{jk}$ $ \forall$ $ k \in \{1, \cdots, K\} $. Therefore it is sufficient to normalise each $\Delta_{jk}$ as 
\begin{equation}
r_{jk} = \frac{\Delta_{jk}}{\sum_{h=1}^{K}\Delta_{jh}} 
\end{equation}

\begin{equation}
\begin{split}
 q^*(\pmb z | X) &= \prod_j \prod_k  r_{jk}^{z_{jk}}\\
 &=  \prod_j q^*(z_j)\\
\end{split}
\end{equation}

The expectation for the discrete distribution $q^*(z_{jk})$ gives the responsibilities $r_{jk}$ for point $x_j$ with the current $k^{th}$ cluster's parameters :
\begin{equation}
\begin{split}
 \mathbb{E}_{q^*(z_{jk})}[z_{jk}]  &:=\frac{\pi_k \mathcal{N}(\pmb x_j|\alpha_j \pmb \mu_k, \beta_j \Sigma_k)}{\sum_{m=1}^K \pi_m \mathcal{N}(\pmb x_j|\alpha_j \pmb \mu_m, \beta_j \Sigma_m)} = r_{jk} \\
\end{split}
\end{equation}
with the overall runtime complexity for calculating $r_{jk}$ is $\mathrm{O}(d^3)$.
\paragraph{MAP estimate for z}
Since the E-step has a runtime $\sim \mathrm{O}(d^3)$ due to three matrix inversions, we substitute this as $z_{MAP}(\pmb x) = \arg\max_z p(\pmb x|\pmb z)p(\pmb z|\pmb \pi)$ which has a runtime of $ \sim \mathrm{O}(d^2)$ when $\Sigma_k^{-1}$s and $\Sigma^{\prime \prime}$ are apriori Cholesky decomposed.


\paragraph{Variational M-step.}
Take the expectation of the log of the joint distribution with respect to $\pmb z$. We have:

\begin{equation}
\begin{split}
\ln q^*( & \pmb \pi,\pmb \mu, \Sigma,\pmb \mu^\prime, \Sigma^\prime, R, S, M , \lambda, \pmb \alpha, \pmb \beta ,\pmb{p},\eta,\gamma,\Lambda,\zeta|X,C) \\&= \mathbb{E}_{\pmb z}[\ln p(X,C, \pmb z, \pmb \pi,\pmb \mu, \Sigma,\pmb \mu^\prime, \Sigma^\prime, R, S, M , \lambda, \pmb \alpha, \pmb \beta ,\pmb{p},\eta,\gamma,\Lambda, \zeta )] + C\\
 &= \mathbb{E}_{\pmb z}[ \ln p(X|\pmb z,\pmb \mu, \Sigma,\pmb \alpha, \pmb \beta) + \ln p(C|\pmb p, \pmb \pi, \zeta) + \ln p(\pmb z|\pmb \pi)+ \ln p(\pmb \pi|\pmb \varphi) +\\
&  \ln p(\pmb \mu^\prime|\pmb \mu^{\prime\prime}, \Sigma^{\prime\prime}) + \ln p(\Sigma^{-1\prime}|\Sigma^{-1\prime\prime}, d) +\\
&  \ln p(\pmb \alpha| \nu, \delta) + \ln p(\pmb \beta|\omega, \theta)] + \\
& \sum_k \Big( \ln p(\pmb \mu_k|\pmb \mu^\prime, \Sigma^\prime)+ \ln p(\Sigma^{-1  }_k|\pmb R_k,\gamma)+ \ln p(\pmb R_k|SM\pmb{p_k}, \lambda)+ \ln  p(\pmb p_k| \eta, \Lambda) \Big)] +Const
\end{split}
\end{equation}

Taking terms in $ (\pmb \pi,\pmb \mu, \Sigma,\pmb \mu^\prime, \Sigma^\prime, R, S, M , \lambda, \pmb \alpha, \pmb \beta ,\pmb{p},\eta,\gamma,\Lambda, \zeta)$, we get:
\begin{equation}
\begin{split}
\ln & q^*( \pmb \pi,\pmb \mu, \Sigma,\pmb \mu^\prime, \Sigma^\prime, R, S, M , \lambda, \pmb \alpha, \pmb \beta ,\pmb{p},\eta,\gamma,\Lambda |X, C) 
 = \\ &\mathbb{E}_{\pmb z}[ \ln p(X|\pmb z,\pmb \mu, \Sigma,\pmb \alpha, \pmb \beta) +  \\
 & \ln p(\pmb z|\pmb \pi)+ \ln p(C|\pmb p, \pmb \pi, \zeta) + \ln p(\pmb \pi|\pmb \varphi) +\\
&  \ln p(\pmb \mu^\prime|\pmb \mu^{\prime\prime}, \Sigma^{\prime\prime}) + \ln p(\Sigma^{-1\prime}|\Sigma^{-1\prime\prime}, d) +
  \ln p(\pmb \alpha| \nu, \delta) + \ln p(\pmb \beta|\omega, \theta)] + \\
& \sum_k \Big( \ln p(\pmb \mu_k|\pmb \mu^\prime, \Sigma^\prime)+ \ln p(\Sigma^{-1  }_k| (R_k+R_k^T)^2,\gamma)+ \ln p( R_k|SM\pmb{p_k}, \lambda)+ \ln  p(\pmb p_k| \eta, \Lambda) \Big)]\\ & +C\\
&= \mathbb{E}_{\pmb z}\Big[ \ln \prod_j \prod_k \mathcal{N}(\pmb x_j|\alpha_j \pmb \mu_k, \beta_j \Sigma_k)^{z_{jk}} + \ln \prod_j \prod_k  \pi_k^{z_{jk}}\Big] + \ln \mathrm{Dir} (\pmb \pi | \varphi_0) +\\
&  \ln \prod_t^r \prod_k \mathcal{N}(c_t|\sum_k \pi_k \pmb p_k,\zeta\mathrm{I}) +    \ln \mathcal{N}(\pmb \mu^\prime | \pmb \mu^{\prime\prime},\Sigma^{\prime\prime}) + \ln \mathcal{W}( \Sigma^{\prime -1} | \Sigma^{\prime \prime -1},d) +\\
& \ln \prod_j \mathrm{logNormal}( \alpha_j|\nu,\delta) + \ln \prod_j \mathrm{logNormal}( \beta_j|\omega,\theta) + \\
& \ln \Big( \prod_k \mathcal{N}(\pmb \mu_k|\pmb \mu^\prime, \Sigma^\prime) \mathrm{Wishart}( \Sigma^{-1  }_k| R_k,\gamma) 
 \mathcal{N}( (R_k+R_k^T)^2| S  M\pmb{p}_k, \lambda) \mathrm{trunc}\mathcal{N}(\pmb p_k| \eta, \Lambda) \Big)  \\
 & + Const \\
 &= \mathbb{E}_{\pmb z}\Big[  \sum_j \sum_k z_{jk} \ln \mathcal{N}(\pmb x_j|\alpha_j \pmb \mu_k, \beta_j \Sigma_k) + \sum_j \sum_k z_{jk} \ln \pi_k \Big] + 
 \ln \mathrm{Dir} (\pmb \pi | \varphi_0) + \\
 &  \sum_t^r \sum_k \ln \mathcal{N}(c_t|\sum_k \pi_k \pmb p_k,\zeta\mathrm{I}) +  \ln \mathcal{N}(\pmb \mu^\prime | \pmb \mu^{\prime\prime},\Sigma^{\prime\prime}) + \ln \mathcal{W}( \Sigma^{\prime -1} | \Sigma^{\prime \prime -1},d) +\\
&  \sum_j \ln \mathrm{logNormal}( \alpha_j|\nu,\delta) +  \sum_j \ln \mathrm{logNormal}( \beta_j|\omega,\theta) + \\
&  \sum_k \ln \mathcal{N}(\pmb \mu_k|\pmb \mu^\prime, \Sigma^\prime) + \sum_k \ln  \mathrm{Wishart}( \Sigma^{-1  }_k| (R_k+R_k^T)^2,\gamma) + \\
& \sum_k \ln 
 \mathcal{N}( R_k| S  M\pmb{p}_k, \lambda) + \sum_k \ln  \mathrm{trunc}\mathcal{N}(\pmb p_k| \eta, \Lambda)   + Const \\
 %
\end{split}
\label{eq: Mstep_2}
\end{equation}


We assume that the set of latent variables is independent of the rest of the latent variables given $X$ and $C$. This independence assumption reduces the problem complexity and allows us to get closed-form solutions in the M-step. This is called the \emph{mean-field assumption.}
We use this assumption and proceed to factor the latent variables into conditionally-independent components, to perform co-ordinate ascent mean field VI (CAVI) on each variational component:
\begin{equation}
\begin{split}
\ln & q^*( \pmb \pi,\pmb \mu, \Sigma,\pmb \mu^\prime, \Sigma^\prime, R, S, M , \lambda, \pmb \alpha, \pmb \beta ,\pmb{p},\eta,\gamma,\Lambda, \zeta |X, C)  \\
& = \ln q^*(\pmb \pi) + \ln q^*(\pmb \mu) + \ln q^*(\Sigma) + \ln q^*(\pmb \alpha)  + \ln q^*(\pmb \beta)  \\ & + \ln q^*(\pmb \mu^\prime) +  + \ln q^*(\Sigma^\prime)  + \ln q^*( R) + \ln q^*(\pmb p)\\
&=\ln q^*(\pmb \pi) + \ln \prod_k q^*(\mu_k) + \ln \prod_k q^*(\Sigma_k^{-1}) + \ln \prod_j q^*(\alpha_j)  + \ln \prod_j q^*(\beta_j)  \\ & + \ln q^*(\pmb \mu^\prime) +  + \ln q^*(\Sigma^\prime)  + \ln \prod_k q^*( R) + \ln \prod_k q^*(\pmb p)\\
&=\ln q^*(\pmb \pi) +  \sum_k \ln q^*(\mu_k) + \sum_k \ln q^*(\Sigma_k^{-1}) +  \sum_j \ln q^*(\alpha_j)  +  \sum_j \ln q^*(\beta_j)  \\ & + \ln q^*(\pmb \mu^\prime) +  + \ln q^*(\Sigma^\prime)  + \sum_k \ln q^*( R) + \sum_k \ln q^*(\pmb p)\\
\end{split}
\label{eq: Mstep_3}
\end{equation}

Let us find the approximate distributions $q^*(\cdot)$ for every parameter in the RHS of Equation \ref{eq: Mstep_3} by comparing to RHS of Equation \ref{eq: Mstep_2}.

\begin{enumerate}

\item
\begin{equation}
\begin{split}
q^*(\pmb \pi) = p(\pmb \pi| \varphi) &=\mathrm{Stick} (\pmb \pi | \varphi) \\
p( \pi_k^\prime| \varphi)&=\mathrm{Beta} (\pi_k^\prime | 1, \varphi) \\
 \pi_k &= \pi_k^\prime \prod_i^{K-1}(1-\pi_i^\prime) \\
\end{split}
\end{equation}

\item 
\begin{equation}
\begin{split}
\sum_k \ln q^*(\mu_k) &= \mathbb{E}_{\pmb z} \Big[ \sum_j \sum_k {z_{jk}} \ln \mathcal{N}(\pmb x_j|\alpha_j\pmb\mu_k, \beta_j \Sigma_k) \Big] + \sum_k \ln \mathcal{N}(\pmb \mu_k |\pmb \mu^\prime,\Sigma^\prime)\\
 &= \sum_k \underbrace{(\mathbb{E}_{\pmb z} \Big[ \sum_j {z_{jk}} \ln \mathcal{N}(\pmb x_j|\alpha_j \pmb \mu_k, \beta_j \Sigma_k) \Big] +  \ln \mathcal{N}(\pmb \mu_k | \pmb \mu^\prime,\Sigma^\prime))}_{expanded \,\, below}\\
 &=\sum_j  \mathbb{E}_{\pmb z}  {z_{jk}} \ln \mathcal{N}(\pmb x_j|\alpha_j \pmb \mu_k, \beta_j \Sigma_k) +  \ln \mathcal{N}(\pmb \mu_k | \pmb \mu^\prime,\Sigma^\prime)\\
&= -\frac{1}{2}\sum_j  \mathbb{E}_{\pmb z}  {z_{jk}}( \pmb \mu_k - \frac{\pmb x_j}{\alpha_j})^T {(\frac{\beta_j}{\alpha_j^2} \Sigma_k)}^{-1}(\pmb \mu_k - \frac{x_j}{\alpha_j}) \\
& - \frac{1}{2}\sum_j  \mathbb{E}_{\pmb z}  {z_{jk}}\ln|\beta_j\Sigma_k| -\frac{d}{2}\sum_j  \mathbb{E}_{\pmb z}  {z_{jk}}\ln (2\pi)\\
&-\frac{1}{2}\ln |\Sigma^\prime|-\frac{d}{2} \ln (2\pi)-\frac{1}{2}(\pmb\mu_k - \pmb\mu^\prime)^T \Sigma^{\prime -1}(\pmb\mu_k - \pmb\mu^\prime) \\ 
&\propto -\frac{1}{2}\sum_j  \mathbb{E}_{\pmb z} [ {z_{jk}}( \pmb\mu_k - \frac{\pmb x_j}{\alpha_j})^T {(\frac{\beta_j}{\alpha_j^2} \Sigma_k)}^{-1}(\pmb \mu_k - \frac{\pmb x_j}{\alpha_j}) ]
-\frac{1}{2}(\pmb \mu_k - \pmb \mu^\prime)^T \Sigma^{\prime -1}(\pmb \mu_k - \pmb \mu^\prime) \\ 
& \text{(by taking terms in $\mu_k$ and $\mu^{\prime}$)}\\
&\propto -\frac{1}{2}\sum_j  \mathbb{E}_{\pmb z}  {z_{jk}}\mathbb{E}_{\pmb z}[( \pmb \mu_k - \frac{\pmb x_j}{\alpha_j})^T {(\frac{\beta_j}{\alpha_j^2} \Sigma_k)}^{-1}(\pmb \mu_k - \frac{\pmb x_j}{\alpha_j})] 
-\frac{1}{2}(\pmb \mu_k - \pmb \mu^\prime)^T \Sigma^{\prime -1}(\pmb \mu_k - \pmb \mu^\prime) \\ 
&\propto -\frac{1}{2}\sum_j  \mathbb{E}_{\pmb z}  {z_{jk}} \Big(\textit{trace(}\Sigma_k^{-1}(\tfrac{\beta_j}{\alpha_j^2})^{-1}\Sigma^{\prime \prime -1}) +( \pmb{\bar \mu_k} - \frac{\pmb x_j}{\alpha_j})^T {(\frac{\beta_j}{\alpha_j^2} \Sigma_k)}^{-1}(\pmb {\bar \mu_k }- \frac{\pmb  x_j}{\alpha_j})\Big) \\
& -\frac{1}{2}(\pmb \mu_k - \pmb\mu^\prime)^T \Sigma^{\prime -1}(\pmb\mu_k - \pmb\mu^\prime) \\ 
&\propto -\frac{1}{2} \sum_k \Big(\sum_j  r_{jk}  \Big(\textit{trace}(\Sigma_k^{-1}(\tfrac{\beta_j}{\alpha_j^2})^{-1}\Sigma^{\prime \prime -1}) +( \pmb{\bar  \mu_k} - \frac{\pmb x_j}{\alpha_j})^T {(\frac{\beta_j}{\alpha_j^2} \Sigma_k)}^{-1}(\pmb{\bar  \mu_k }- \frac{\pmb x_j}{\alpha_j})\Big) \\
& + (\pmb \mu_k - \pmb {\mu}^\prime)^T \Sigma^{\prime -1}(\pmb \mu_k - \pmb {\mu}^\prime) \Big) \\ 
 \end{split}
\label{eq: Mstep_muk}
\end{equation}

\item Let us denote $R^*_k = (R_k + R_k^T)^2$. By construction of $R_k$, $R^*_k$ is a Gram matrix and a valid scale matrix for Wishart parametrisation. 
\begin{equation}
\begin{split}
\sum_k \ln q^*(\Sigma_k^{-1}) &= \mathbb{E}_{\pmb z} \Big[ \sum_j \sum_k {z_{jk}} \ln \mathcal{N}(\pmb  x_j|\alpha_j \pmb  \mu_k, \beta_j \Sigma_k) \Big] + \sum_k \ln \mathcal{W}(  \Sigma^{-1}_k | R^*_k,\lambda)\\
&= \sum_k \Big[ \underbrace{\sum_j \mathbb{E}_{\pmb z} {z_{jk}} \ln \mathcal{N}(\pmb  x_j|\alpha_j \pmb  \mu_k, \beta_j \Sigma_k) + \ln \mathcal{W}(  \Sigma^{-1}_k |R^*_k,\lambda)}_{expanded \,\, below} \Big] \\
 &=  -\sum_j \mathbb{E}_{\pmb z} {z_{jk}} \frac{d}{2} \ln 2\pi - \sum_j \mathbb{E}_{\pmb z} {z_{jk}} \frac{1}{2} \ln \beta_j^d- \\
& \sum_j \mathbb{E}_{\pmb z} {z_{jk}} \frac{1}{2}\ln |\Sigma_k| -\sum_j \mathbb{E}_{\pmb z} {z_{jk}}  \frac{1}{2\beta_j}(\pmb  x_j-\alpha_j\pmb  \mu_k)^T(\Sigma_k)^{-1}(\pmb  x_j-\alpha_j\pmb  \mu_k)+ \\
& \ln |\Sigma_k^{-1}|^{\frac{\lambda-d-1}{2}} + \{-\frac{tr(R^{*-1}_k\Sigma_k^{-1})}{2}\} - \ln(2^{\lambda d/2} |R^{*-1}_k|^{-\lambda/2} \Gamma_d(\lambda/2))  \\ 
&\propto -\sum_j \mathbb{E}_z z_{jk}(\tfrac{1}{2}\ln \beta_j^d + \tfrac{1}{2}\ln |\Sigma_k|) \\ &-\sum_j \mathbb{E}_z z_{jk} \tfrac{1}{2\beta_j} \Big(\textit{trace}(\Sigma_k^{-1}\Sigma^{\prime \prime -1}) + (\pmb \mu^{\prime \prime} - \alpha_j\pmb  \mu_k)^T  (\Sigma_k)^{-1} (\pmb  \mu^{\prime \prime} - \alpha_j\pmb \mu_k)  \Big) \\
& + \ln |\Sigma_k^{-1}|^{\frac{\lambda-d-1}{2}} + \{-\frac{tr(R^{*-1}_k\Sigma_k^{-1})}{2}\} - \ln(2^{\lambda d/2} |R^{*-1}_k|^{-\lambda/2} \Gamma_d(\lambda/2))  \\ 
\end{split}
\label{eq: Mstep_Sigmak}
\end{equation}

\item If $\alpha_j \sim \log\mathcal{N}(\nu, \delta)$, then $\alpha_j^* = \ln \alpha_j \sim \mathcal{N}(\nu, \delta)$ by properties of log Normal distribution \cite{aitchison1957lognormal}.

\begin{equation}
\begin{split}
\sum_j \ln q^*(\alpha_j) &= \mathbb{E}_{\pmb z} \Big[ \sum_j \sum_k {z_{jk}} \ln \mathcal{N}(\pmb  x_j|\alpha_j \pmb  \mu_k, \beta_j \Sigma_k) \Big] +\sum_j \ln \mathcal{N}( \alpha_j|\nu,\delta)\\
&= \mathbb{E}_{\pmb z} \Big[ \sum_j \sum_k {z_{jk}} \ln \mathcal{N}(\pmb x_j|\alpha_j \pmb  \mu_k, \beta_j \Sigma_k) \Big] +\sum_j \mathcal{N}( \alpha_j^*|\nu,\delta)\\
&=\sum_j \underbrace{( \Big[  \sum_k \mathbb{E}_{\pmb z}{z_{jk}} \ln \mathcal{N}(\pmb  x_j|\alpha_j \pmb  \mu_k, \beta_j \Sigma_k) \Big] + \mathcal{N}( \alpha_j^*|\nu,\delta))}_{expanded \,\, below}\\
& = - \frac{1}{2}\sum_k  \mathbb{E}_{\pmb z}  {z_{jk}}\ln|\beta_j\Sigma_k| -\frac{d}{2}\sum_k \mathbb{E}_{\pmb z}  {z_{jk}}\ln (2\pi) - \\ &\sum_k  \mathbb{E}_{\pmb z}{z_{jk}} \frac{1}{2}(\pmb  x_j - \alpha_j \pmb  \mu_k)^T {(\beta_j \Sigma_k)}^{-1}(\pmb x_j - \alpha_j \pmb  \mu_k) +\\
&  \frac{1}{\sqrt{2\delta^2\pi}} \exp \Big(- \frac{(\alpha_j^*-\nu)\mathbb{I}_{1\times 1}(\alpha_j^*-\nu)}{2\delta^2} \Big)\\
& \propto -\sum_k  \mathbb{E}_{\pmb z}{z_{jk}} \frac{1}{2}(\pmb x_j - \alpha_j \pmb \mu_k)^T {(\beta_j \Sigma_k)}^{-1}(\pmb x_j - \alpha_j \pmb \mu_k) + \\ &
\frac{1}{\sqrt{2\delta^2\pi}} \exp \Big(- \frac{(\alpha_j^*-\nu)\mathbb{I}_{1\times 1}(\alpha_j^*-\nu)}{2\delta^2} \Big) \,\, \text{(taking terms in $\alpha_j$)}\\
& \propto -\sum_k  r_{jk} \frac{1}{2}(\textit{trace}(\Sigma_k^{-1}\beta^{-1}\Sigma^{\prime \prime -1}) + (\pmb \mu^{\prime \prime} - \alpha_j\pmb \mu_k)^T  (\beta\Sigma_k)^{-1} (\pmb  \mu^{\prime \prime} - \alpha_j\pmb \mu_k)) + \\ 
& \frac{\alpha_j}{\alpha_j \sqrt{2\delta^2\pi}} \exp \Big(- \frac{(\ln \alpha_j -\nu)\mathbb{I}_{1\times 1}(\ln \alpha_j-\nu)}{2\delta^2} \Big) \,\, \\ & \text{(by replacing $\alpha_j^*$ with $\ln \alpha_j$ and making this a logNormal pdf )}\\
\end{split}
\label{eq: Mstep_alphaj}
\end{equation}


\item Derivation is similar to that of Equation \ref{eq: Mstep_alphaj}.
\begin{equation}
\begin{split}
\sum_j \ln q^*(\beta_j) &= \mathbb{E}_{\pmb z} \Big[ \sum_j \sum_k {z_{jk}} \ln \mathcal{N}(\pmb x_j|\alpha_j \pmb \mu_k, \beta_j \Sigma_k) \Big] +\sum_j \ln \mathrm{log}\mathcal{N}( \beta_j|\omega,\theta)\\
 &= \mathbb{E}_{\pmb z} \Big[ \sum_j \sum_k {z_{jk}} \ln \mathcal{N}(\pmb x_j|\alpha_j \pmb \mu_k, \beta_j \Sigma_k) \Big] +\sum_j \mathcal{N}( \beta_j^*|\omega,\theta)\\
& \propto -\sum_k  r_{jk} \frac{1}{2}(\textit{trace}(\Sigma_k^{-1}\beta^{-1}\Sigma^{\prime \prime -1}) + (\pmb \mu^{\prime \prime} - \alpha_j\pmb \mu_k)^T  (\beta\Sigma_k)^{-1} (\pmb \mu^{\prime \prime} - \alpha_j\pmb \mu_k)) + \\ 
& \frac{\beta_j}{\beta_j \sqrt{2\theta^2\pi}} \exp \Big(- \frac{(\ln \beta_j -\omega)\mathbb{I}_{1\times 1}(\ln \beta_j - \omega)}{2\theta^2} \Big) \,\, \\ & \text{(by replacing $\beta_j^*$ with $\ln \beta_j$ and making this a logNormal pdf )}\\
\end{split}
\label{eq: Mstep_betaj}
\end{equation}

\item
\begin{equation}
\begin{split}
\ln q^*(\pmb \mu^\prime) &=  \ln \mathcal{N}(\pmb \mu^\prime | \pmb \mu^{\prime\prime},\Sigma^{\prime\prime}) + \sum_k \ln \mathcal{N}(\pmb \mu_k | \pmb \mu^\prime,\Sigma^\prime) \\
&\sim \ln \mathcal{N}(\pmb \mu_{\mu^\prime},\Sigma_{\mu^\prime})\\
\pmb \mu_{\mu^\prime} &= \Sigma_{\mu^\prime} (\Sigma^{\prime\prime-1} \pmb \mu^{\prime\prime} + K^2\Sigma^{\prime-1}\pmb {\bar{\mu^\prime}})\\
\Sigma_{\mu^\prime} &= (\Sigma^{\prime\prime-1} + K\Sigma^{\prime-1})^{-1}\\
\end{split}
\label{eq: Mstep_muprime}
\end{equation}

\item 
\begin{equation}
\begin{split}
\ln q^*(\Sigma^{-1 \prime})  &= \ln \mathcal{W}( \Sigma^{\prime -1} | \Sigma^{\prime \prime -1},d) +  \sum_k \ln \mathcal{N}(\pmb \mu_k | \pmb \mu^\prime,\Sigma^\prime)\\
&\sim \ln \mathcal{W}(V_{\Sigma^{\prime-1}},d_{\Sigma^{\prime-1}})\\
V_{\Sigma^{\prime-1}} &= (d\Sigma^{\prime\prime}+ 2\Sigma_{rss})^{-1}\\
d_{\Sigma^{\prime-1}} &= d + K \\\end{split}
\label{eq: Mstep_Sigmaprime}
\end{equation}

\item 
\begin{equation}
\begin{split}
\sum_k \ln q^*( R_k)  &= \mathbb{E}_{(\pi_k,\pmb  p_k)}[\sum_t^r \sum_k \ln \mathcal{N}(c_t| \sum_k(\pi_k p_k), \zeta\mathbb{I})] + \sum_k\ln \mathcal{W}(\Sigma_k^{-1}|R_k^*, \gamma) + \\ & \sum_k \sum_i \sum_{i^\prime} \ln  \mathcal{N}( R_k^{i,i^\prime}| S^{i,i^\prime}  M^{i,i^\prime} \pmb p_k^{g(i,i^\prime)},\lambda)\\
 \ln q^*( R_k)  &= \mathbb{E}_{(\pi_k,\pmb p_k)}[\sum_t^r  \ln \mathcal{N}(c_t| \sum_k(\pi_k \pmb  p_k), \zeta\mathbb{I})] +\ln \mathcal{W}(\Sigma_k^{-1}|R_k^*, \gamma) + \\ & \sum_i \sum_{i^\prime} \ln  \mathcal{N}( R_k^{i,i^\prime}| S^{i,i^\prime}  M^{i,i^\prime} \pmb p_k^{g(i,i^\prime)},\lambda)\\
    &= \sum_t^r \mathbb{E}_{(\pi_k,\pmb p_k)}[  (c_t - \sum_k(\pi_k \pmb p_k))^T (\zeta\mathbb{I})^{-1} (c_t - \sum_k(\pi_k\pmb  p_k))] +\ln \mathcal{W}(\Sigma_k^{-1}|R_k^*, \gamma) + \\ & \sum_i \sum_{i^\prime} \ln  \mathcal{N}( R_k^{i,i^\prime}| S^{i,i^\prime}  M^{i,i^\prime} \pmb p_k^{g(i,i^\prime)},\lambda)\\
& \propto \sum_t^r \textit{trace}(\zeta\mathbb{I})^{-1} +  (\bar c_t - \sum_k(\pi_k \pmb p_k))^T (\zeta\mathbb{I})^{-1} (\bar c_t - \sum_k(\pi_k \pmb p_k))] + \\ & \{-\frac{tr(R^{*-1}_k\Sigma_k^{-1})}{2}\} - \ln(2^{\lambda d/2} |R^{*-1}_k|^{-\lambda/2} \Gamma_d(\lambda/2))   \\ & -\tfrac{1}{2\delta^2} \sum_i \sum_{i^\prime} \Big( (R_k^{i,i^\prime}- S^{i,i^\prime}  M^{i,i^\prime} \pmb p_k^{g(i,i^\prime)})\mathbb{I}_{1\times 1}(R_k^{i,i^\prime}- S^{i,i^\prime}  M^{i,i^\prime} \pmb p_k^{g(i,i^\prime)}) \Big)\\
& \propto  \{-\frac{tr(R^{*-1}_k\Sigma_k^{-1})}{2}\} - \ln(2^{\lambda d/2} |R^{*-1}_k|^{-\lambda/2} \Gamma_d(\lambda/2))   \\ & -\tfrac{1}{2\delta^2} \sum_i \sum_{i^\prime} \Big( (R_k^{i,i^\prime}- S^{i,i^\prime}  M^{i,i^\prime} \pmb p_k^{g(i,i^\prime)})\mathbb{I}_{1\times 1}(R_k^{i,i^\prime}- S^{i,i^\prime}  M^{i,i^\prime} \pmb p_k^{g(i,i^\prime)}) \Big) \quad \\& \text{(taking terms only in $R_k$)}\\
\end{split}
\label{eq: Mstep_R}
\end{equation}

\item 
\begin{equation}
\begin{split}
\sum_k \ln q^*(\pmb p_k)  &= \mathbb{E}_{(\pi_k,\pmb p_k)}[\sum_t^r \sum_k \ln \mathcal{N}(c_t| \sum_k(\pi_k \pmb p_k), \zeta\mathbb{I})] + \sum_k\ln \mathrm{trunc}\mathcal{N}(\pmb p_k|\eta, \Lambda) + \\& \sum_k \sum_i \sum_{i^\prime} \ln  \mathcal{N}( R_k^{i,i^\prime}| S^{i,i^\prime}  M^{i,i^\prime} \pmb p_k^{g(i,i^\prime)},\lambda)\\
 \ln q^*(\pmb p_k)  &= \mathbb{E}_{(\pi_k,\pmb p_k)}[\sum_t^r  \ln \mathcal{N}(c_t| \sum_k(\pi_k \pmb p_k), \zeta\mathbb{I})] + \ln \mathrm{trunc}\mathcal{N}(\pmb p_k|\eta, \Lambda) +\\&  \sum_i \sum_{i^\prime} \ln  \mathcal{N}( R_k^{i,i^\prime}| S^{i,i^\prime}  M^{i,i^\prime} \pmb p_k^{g(i,i^\prime)},\lambda)\\
 &= \sum_t^r \mathbb{E}_{(\pi_k,\pmb p_k)}[  (c_t - \sum_k(\pi_k \pmb p_k))^T (\zeta\mathbb{I})^{-1} (c_t - \sum_k(\pi_k \pmb p_k))] +\ln \Big(\tfrac{1}{\sqrt{2\pi}\Lambda}\exp(-\tfrac{1}{2} (\tfrac{\pmb p_k-\eta}{\Lambda})^2))\Big)  \\ & -\tfrac{1}{2}\ln\Big( (1+ \textit{erf}(\tfrac{\infty - \eta}{\Lambda\sqrt{2}})) - (1+ \textit{erf}(\tfrac{ - \eta}{\Lambda\sqrt{2}}))\Big) + \sum_i \sum_{i^\prime} \ln  \mathcal{N}( R_k^{i,i^\prime}| S^{i,i^\prime}  M^{i,i^\prime} \pmb p_k^{g(i,i^\prime)},\lambda)\\
 & \propto \sum_t^r \textit{trace}(\zeta\mathbb{I})^{-1} +  (\bar c_t - \sum_k(\pi_k p_k))^T (\zeta\mathbb{I})^{-1} (\bar c_t - \sum_k(\pi_k p_k))] + \ln \Big(\tfrac{1}{\sqrt{2\pi}\Lambda}\exp(-\tfrac{1}{2} (\tfrac{\pmb p_k-\eta}{\Lambda})^2))\Big)   \\ & -\tfrac{1}{2\delta^2} \sum_i \sum_{i^\prime} \Big( (R_k^{i,i^\prime}- S^{i,i^\prime}  M^{i,i^\prime} \pmb p_k^{g(i,i^\prime)})\mathbb{I}_{1\times 1}(R_k^{i,i^\prime}- S^{i,i^\prime}  M^{i,i^\prime} \pmb p_k^{g(i,i^\prime)}) \Big) \quad \\& \text{(taking terms only in $\pmb p_k$)}\\
\end{split}
\label{eq: Mstep_p}
\end{equation}

\end{enumerate}


\section{Blueprint for Variational algorithm}
\begin{enumerate}
\item \textbf{Perform} Variational E-step 
\subitem a. Compute $q^{*}(z_n) = \prod_k r_{jk}^{z_{jk}}$ where 
\begin{equation}
r_{jk} \propto  \widetilde{|\beta_j \Sigma_k|^{-1}} \exp (-S_2)  \widetilde\pi_k,
\end{equation}
$\sum_k r_{nk} =1$ and $S_2 = \tfrac{1}{2}\Big(\textit{trace}(\Sigma_k^{-1}\beta_j^{-1}\Sigma^{\prime \prime -1}) + (\pmb \mu^{\prime \prime} - \alpha_j\pmb \mu_k)^T  (\beta_j\Sigma_k)^{-1} (\pmb \mu^{\prime \prime} - \alpha_j\pmb \mu_k) \Big)$

\item \textbf{Perform} Variational M-step
\subitem b. Compute  $q^{*}(\pi_k) = \pi_k \sim \text{Stick-breaking Beta}(1,\varphi)$  

\subitem c. Compute  $q^{*}(\mu_k) =  \exp \Big( -\tfrac{1}{2} \sum_j  r_{jk}  \Big(\textit{trace}(\Sigma_k^{-1}(\tfrac{\beta_j}{\alpha_j^2})^{-1}\Sigma^{\prime \prime -1}) +( \pmb {\bar \mu_k} - \frac{\pmb x_j}{\alpha_j})^T {(\frac{\beta_j}{\alpha_j^2} \Sigma_k)}^{-1}(\pmb {\bar \mu_k} - \frac{\pmb x_j}{\alpha_j})
+ (\pmb \mu_k - \pmb \mu^\prime)^T \Sigma^{\prime -1}(\pmb \mu_k - \pmb \mu^\prime) \Big) + const \Big) $ from Equation \ref{eq: Mstep_muk}.

\subitem d. Compute $q^{*}(\Sigma_k^{-1}) = \exp \Big( -\tfrac{1}{2}\sum_j r_{jk}\Big(\ln \beta_j^d + \ln |\Sigma_k| + \tfrac{1}{\beta_j} \Big(\textit{trace}(\Sigma_k^{-1}\Sigma^{\prime \prime -1}) + (\pmb \mu^{\prime \prime} - \alpha_j\pmb \mu_k)^T  (\Sigma_k)^{-1} (\pmb \mu^{\prime \prime} - \alpha_j\pmb \mu_k)  \Big) \Big)  + \ln |\Sigma_k^{-1}|^{\frac{\lambda-d-1}{2}} + \{-\frac{tr(R^{*-1}_k\Sigma_k^{-1})}{2}\} - \ln(2^{\lambda d/2} |R^{*-1}_k|^{-\lambda/2} \Gamma_d(\lambda/2)) +const \Big) $  
from Equation \ref{eq: Mstep_Sigmak}.


\subitem e. Compute  $q^{*}(\alpha_j) = \exp \Big(-\sum_k  r_{jk} S_2 + 
 \frac{1}{ \sqrt{2\delta^2\pi}} \exp \Big(- \frac{(\ln \alpha_j -\nu)\mathbb{I}_{1\times 1}(\ln \alpha_j-\nu)}{2\delta^2} \Big) + const \Big)$ from Equation \ref{eq: Mstep_alphaj}.


\subitem f. Compute  $q^{*}(\beta_j) = \exp\Big(-\sum_k  r_{jk} S_2 +
 \frac{1}{\sqrt{2\theta^2\pi}} \exp \Big(- \frac{(\ln \beta_j -\omega)\mathbb{I}_{1\times 1}(\ln \beta_j - \omega)}{2\theta^2} \Big)+ const \Big) $ from Equation \ref{eq: Mstep_betaj}.

\subitem g. Compute $q^*(\pmb \mu^\prime) \sim  \mathcal{N}(\mu_{\mu^\prime},\Sigma_{\mu^\prime}) $ from Equation \ref{eq: Mstep_muprime}. 

\subitem h. Compute  $q^*(\Sigma^{-1 \prime}) \sim  \mathcal{W}(V_{\Sigma^{\prime-1}},d_{\Sigma^{\prime-1}}) $ from Equation \ref{eq: Mstep_Sigmaprime}. 

\subitem i. Compute $q^*(R_k) = \exp \Big( \{-\frac{tr(R^{*-1}_k\Sigma_k^{-1})}{2}\} - \ln(2^{\lambda d/2} |R^{*-1}_k|^{-\lambda/2} \Gamma_d(\lambda/2))    -\tfrac{1}{2\delta^2} \sum_i \sum_{i^\prime} \Big( (R_k^{i,i^\prime}- S^{i,i^\prime}  M^{i,i^\prime} \pmb p_k^{g(i,i^\prime)})\mathbb{I}_{1\times 1}(R_k^{i,i^\prime}- S^{i,i^\prime}  M^{i,i^\prime} \pmb p_k^{g(i,i^\prime)}) \Big)+ const \Big) $

\subitem j. Compute $q^*(\pmb p_k) = \exp \Big(   (\bar c_t - \sum_k(\pi_k \pmb p_k))^T (\zeta\mathbb{I})^{-1} (\bar c_t - \sum_k(\pi_k \pmb p_k))] + \ln \Big(\tfrac{1}{\sqrt{2\pi}\Lambda}\exp(-\tfrac{1}{2} (\tfrac{\pmb p_k-\eta}{\Lambda})^2))\Big)    -\tfrac{1}{2\delta^2} \sum_i \sum_{i^\prime} \Big( (R_k^{i,i^\prime}- S^{i,i^\prime}  M^{i,i^\prime} \pmb p_k^{g(i,i^\prime)})\mathbb{I}_{1\times 1}(R_k^{i,i^\prime}- S^{i,i^\prime}  M^{i,i^\prime} \pmb p_k^{g(i,i^\prime)}) \Big) + const \Big)$

\end{enumerate}

\section{Scalable Implementation}

Given the complexity of the model (refer to Supplementary section C), for a scalable implementation applicable to biological data containing thousands of cells, we used  probabilistic programming languages with several useful approximations and implementation tricks.

\subsection{Stan}

A first implementation of \emph{Symphony} intended for smaller-scale datasets utilizes probabilistic programming language Stan. As Stan uses the No-U-Turn Sampler (NUTS) MCMC algorithm, it produces more accurate results asymptotically and is therefore preferred if computational resources allow. However, since Stan does not support inference for an infinite mixture
model, we simply use a finite mixture model in the experiments
presented herein with the Stan implementation. In addition, we implemented a version with an optional asymmetric, user-defined Dirichlet prior for fair comparison against methods which require prior knowledge of mixture proportions.

\subsection{Edward}

To scale \emph{Symphony} to larger datasets, we implemented the model in probabilistic programming language Edward \cite{tran2016edward}. As Edward is built on a tensorflow back-end, it allows GPU acceleration for faster matrix computation. In addition, the use of variational algorithms allows for faster approximations of the posterior than those which can be obtained with MCMC, albeit with a trade-off in accuracy observed in our case. In order to improve the fit of the model to real data with the Edward implementation, we made a number of approximations below which improved the empirical performance on PBMC and other datasets. 

\subsection{Approximations}
Inference of gene-gene GRNs and covariance matrices are the main goals of \emph{Symphony}, yet accurate inference of such large matrices involves a number of computational challenges. In particular, constraints on covariance matrices of a multivariate normal distribution are difficult to enforce in the optimization setting of variational inference. For instance, large sparse matrices may very easily become non-singular during optimization, leading to un-defined loss functions. 

We use several techniques to solve this problem. For one, we define the Wishart distribution in Edward using the Bartlett Decomposition, rather than the built-in Wishart function of tensorflow, which allows us to more easily define variational parameters. Specifically, we replace the sampling of covariance matrices $\Sigma_k \sim Wishart$ with a generative model constructing only univariate chi-squared distributions $c$ and normal distributions $n$, which can be shown  to produce a valid sample from the Wishart distribution \cite{kshirsagar1959}. In this setting, we define variational distributions corresponding to the dummy variables $n$ and $c$, as opposed to defining a matrix variate distribution which, during the course of optimization, must fit all the constraints of valid covariance matrices. Initialization of these parameters to large values additionally avoids problems with singularity in most cases.

In addition to the use of the Bartlett Decomposition, the Edward version of \emph{Symphony} replaces the standard Wishart with a scaled Wishart for added flexibility of the model in the variational inference case. The scaled Wishart necessitates addition of a latent parameter $\delta$, such that

$$\Sigma_k' \sim Wishart$$
$$\delta_i \overset{\text{iid}}{\sim} Normal$$
$$\Sigma_k = \Delta \Sigma_k' \Delta \text{, where } diag(\Delta)=\delta$$

Addition of the normal distribution above to the generative process infuses flexibility to the Wishart, whose variance is usually defined by a single degrees of freedom parameter. In addition, we allow separate inference of the diagonal and off-diagonal of the covariance matrices. This is a desirable property for \emph{Symphony}, in that the model of gene regulation does not necessarily capture the diagonal of the covariance matrices representing variances of gene expression. Likewise, we solve additional issues caused by matrix inversion by simply replacing the prior on $\Sigma_k$ with a Wishart instead of Inverse-Wishart distribution. We note that, while this choice is not conjugate, this is valid as both distributions satisfy the requirements for priors on the covariance matrix.

We require a variational EM procedure with Edward, such that the cluster assignments $z$ are updated every several iterations with a maximization step. In particular, we choose $z_i$ for each cell based on the maximum likelihood of cluster assignment in the Gaussian mixture. This prevents the need for discrete optimization over categorical variational parameters. The performance of the variational EM algorithm is maximized when a good initial value for clustering is chosen. In this work, we initialized clusters based on cell-cell kNN graphs with Phenograph \cite{levine2015data}.

Finally, we made some other small distributional changes which seemed to produce better results in our experiments with this particular implementation. We replace the multivariate prior on $p$ with a univariate prior centered at a constant. In addition, we treated binary $M$ as a latent variable with a very tight variance. This was to add additional flexibility to the model, and to also assist with singularity issues by providing a dense matrix mean to $R$. We note that all fitted values of $M$ were similar within a small $\epsilon$ to their previously fixed values of either 0 or 1.

\end{document}